\documentclass{IEEEtaes}

\usepackage[colorlinks,urlcolor=blue,linkcolor=blue,citecolor=blue]{hyperref}
\usepackage{color,array}

\usepackage{amsmath,amssymb,amsfonts,bm}
\usepackage{balance}
\usepackage{algorithmic}
\usepackage{algorithm}
\usepackage{array}
\usepackage[caption=false,font=normalsize,labelfont=sf,textfont=sf]{subfig}
\usepackage{textcomp}
\usepackage{stfloats}
\usepackage{url}
\usepackage{verbatim}
\usepackage{graphicx}
\usepackage{cite}
\usepackage{tikz}
\usepackage{booktabs}
\usepackage{multirow}
\usepackage{tikz}
\usepackage{pgfplots} 
\usepackage{makecell}

\usepackage[T1]{fontenc}

\jvol{XX}
\jnum{XX}
\jmonth{XXXXX}
\paper{1234567}
\pubyear{2022}
\doiinfo{TAES.2022.Doi Number}

\begin{document}

\title{Transformer-Based Multi-Object Smoothing with Decoupled Data Association and Smoothing} 

\author{JULIANO PINTO}
\author{GEORG HESS}
\author{YUXUAN XIA}
\member{Member, IEEE}
\author{HENK WYMEERSCH}
\member{Senior Member, IEEE}
\author{LENNART SVENSSON}
\member{Senior Member, IEEE}
\affil{Department of Electrical Engineering, Chalmers University of Technology, 41296 Gothenburg, Sweden.} 

\receiveddate{This work was supported, in part, by a grant from the Chalmers AI Research Centre Consortium. Computational resources were provided by the Swedish National Infrastructure for Computing at C3SE, partially funded by the Swedish Research Council through grant agreement no. 2018-05973.}
%% \accepteddate{XXXXX XX XXXX}
%% \publisheddate{XXXXX XX XXXX}

\corresp{{\itshape (Corresponding author: Juliano Pinto, juliano@chalmers.se)}. }

%\editor{Mentions of supplemental materials and animal/human rights statements can be included here.}
\supplementary{Color versions of one or more of the figures in this article are available online at {http://ieeexplore.ieee.org}.}

\markboth{PINTO ET AL.}{TRANSFORMER-BASED MULTI-OBJECT SMOOTHING WITH DECOUPLED DATA ASSOCIATION AND FILTERING.}
\maketitle

\begin{abstract}
    Multi-object tracking (MOT) is the task of estimating the state trajectories of an unknown and time-varying number of objects over a certain time window. 
    Several algorithms have been proposed to tackle the multi-object smoothing task, where object detections can be conditioned on all the measurements in the time window. However, the best-performing methods suffer from intractable computational complexity and require approximations, performing suboptimally in complex settings. Deep learning based algorithms are a possible venue for tackling this issue but have not been applied extensively in settings where accurate multi-object models are available and measurements are low-dimensional. We propose a novel DL architecture specifically tailored for this setting that decouples the data association task from the smoothing task. We compare the performance of the proposed smoother to the state-of-the-art in different tasks of varying difficulty and provide, to the best of our knowledge, the first comparison between traditional Bayesian trackers and DL trackers in the smoothing problem setting.
\end{abstract}

\begin{IEEEkeywords}
Multi-object smoothing, Deep Learning, Transformers, Random Finite Sets.
\end{IEEEkeywords}

\section{INTRODUCTION}
The objective of multi-object tracking (MOT) is to estimate the state trajectories of an unknown and time-varying number of objects based on a sequence of noisy sensor measurements. The objects can enter and leave the field-of-view (FOV) at any time, there might be missed detections, and there might be false measurements originating from sensor noise and/or clutter. 
%Furthermore, the measurements do not contain information about which was their originating object, requiring algorithms to reason probabilistically about the measurement-object association at any given time-step. 
Being able to achieve high performance in this task is of high importance and has applications in many different contexts, such as pedestrian tracking \cite{mot_in_pedestrian_tracking2}, autonomous driving \cite{mot_in_autonomous_driving1}, sports players tracking \cite{player_tracking}, defense systems \cite{mot_in_military}, animal tracking \cite{mot_for_tracking_animals, fish_tracking}, and more. 

For some applications, the set of trajectories is to be estimated using only measurements available until the most recent time, which can then be used by real-time decision systems downstream (e.g., autonomous driving, defense systems) \cite{mahler2007}. In other applications, all of the measurements in a time window are available when estimating the trajectories (e.g., examining animal behavior recorded in the past, analysis of autonomous driving accidents), and algorithms can condition the estimated trajectories to all of the measurements available, yielding (smoothed) estimates of superior quality \cite{harvey1990forecasting}. This latter formulation will henceforth be referred to as multi-object smoothing.% is considerably harder than single-object smoothing due to the uncertain associations between measurements and objects \cite{mahler2007}.

Several algorithms have been proposed to tackle the multi-object smoothing task. Two important classes are trackers based on approximations of the multi-object density, such as the trajectory Probability Hypothesis Density (PHD) smoothing and trajectory cardinalized PHD smoothing \cite{tphd_and_cphd}, and trackers based on multi-object conjugate priors, such as the generalized labeled multi-Bernoulli \cite{glmb_smoothing}, and the multi-scan trajectory Poisson multi-Bernoulli mixture (TPMBM) \cite{tpmbm_smoothing, granstrom2018poisson, granstrom2019poisson}. Trackers based on multi-object conjugate priors provide closed-form solutions to the task for certain classes of multi-object models. However, due to the unknown measurement-object associations, these approaches suffer from a computational complexity that is super-exponential in the number of time-steps being processed \cite{mahler2007}. This drawback requires them to resort to heuristics such as pruning or merging to maintain computational tractability, which negatively impact performance in challenging scenarios. In addition, if the multi-object models contain nonlinearities, such methods must employ sequential Monte Carlo methods or Gaussian approximations to be able to function, which may further impair tracking quality \cite{sarkka2013bayesian}.

One appealing direction to improve these methods' shortcomings is using deep learning (DL). DL-based algorithms are able to leverage vast amounts of data to learn a variety of complex input-output mappings, such as classifying images \cite{top_imagenet}, translating text \cite{transformers_for_translation}, super-human level of play in classic games \cite{alpha_zero}, and even generating realistic images from text prompts \cite{dall-e2}, so it is plausible that such approaches would be helpful in the MOT context too. Indeed, deep learning has been increasingly applied to the field of MOT, resulting in breakthroughs in state-of-the-art performance \cite{mot_challenge, deep_learning_video_tracking_survey, deep_learning_tracking_survey, chong2021overview}. Initially, DL was used mostly as an aid to solving certain subtasks, such as associating new measurements to existing tracks \cite{chen2017enhancing}, managing track initialization/termination \cite{milan2017online}, and predicting motion models \cite{mot_for_motion_models}, to name a few. More recent approaches use DL to solve the entire (or almost the entire) MOT task, with architectures based on object detector extensions  \cite{tracking_without_bells_and_whistles}, convolutional neural networks \cite{dmm-net, deep_affinity_network, tubetk, fairmot, mots}, graph neural networks \cite{gnn3dmot, braso2020learning, graph_network_for_mot} or, more recently, the transformer network \cite{TrackFormer, motr, transmot, transtrack_mot, memot, segdq}. 

Although such approaches have seen success and widespread adoption in contexts where the measurements are high-dimensional and the multi-object models are too complicated to be modeled accurately (e.g., tracking with video sequences), a setting henceforth referred to as \emph{model-free}, the same cannot be said about the counterpart situation when there are accurate multi-object models available, and the measurements are low-dimensional (e.g., radar tracking of aircraft). In this setting, which we will refer to as the \emph{model-based} setting, random finite sets (RFS) methods based on conjugate priors have been shown to present excellent tracking performance \cite{tpmbm_smoothing}, but with few exceptions \cite{mt3}, there have been no comparisons between the performance of such methods and DL methods based on recent advancements such as the transformer \cite{transformer_paper} architecture, especially for the smoothing task.

This paper proposes a novel DL multi-object tracker specifically tailored for the model-based setting, named Deep Decoupled Data Association and Smoothing (D3AS). Its architecture tackles the smoothing problem formulation by leveraging the recent transformer neural network structure \cite{transformer_paper} for efficient use of the entire measurement sequence when performing predictions. Importantly, it decouples the data association task (reasoning about the unknown association between measurements and objects) from the filtering/smoothing task (predicting trajectories given associations), which reduces training time and model size, and results in more interpretable predictions.
We compare the performance of the proposed tracker to TPMBM 
\cite{granstrom2018poisson, tpmbm_smoothing}, which presents state-of-the-art smoothing performance \cite[Section VIII]{granstrom2019poisson}, in tasks of varying difficulty. To the best of our knowledge, we provide the first results indicating that a deep-learning smoother can outperform model-based Bayesian trackers such as TPMBM in complex tasks. Our specific contributions are:
\begin{itemize}
    \item Propose a novel transformer-based module for the data association task, and the loss formulation for training it independently from the smoothing task. %The resulting model learns to approximate the globally optimal data association given all measurements in a time window. 
    
    \item Propose a novel transformer-based module for the smoothing task, based on recent advancements of large language models \cite{BERT}. In specific, this model takes as input a sequence of measurements and corresponding confidences and predicts a smoothed trajectory.%, using a version of masked language modeling \cite{BERT} to deal with missing measurements.

    \item Perform a thorough comparison between the proposed deep learning smoother and the state-of-the-art smoother TPMBM in various settings of different complexity, reporting that DL methods can outperform traditional model-based models in challenging scenarios.
\end{itemize}

%The rest of this paper is organized as follows. We start by formally defining the multi-object models and the problem formulation in Section \ref{sec:preliminaries}, followed by a short review in section \ref{sec:background} of the transformer encoder architecture, used in the proposed DL-based smoother. Then, in Section \ref{sec:method} we describe the structure of the DL-based smoother, and in Section \ref{sec:evaluation_setting} the evaluation protocol used to compare its performance to TPMBM. This is then followed by Section \ref{sec:results}, describing all of the results of the experimental validation done, and finally a conclusion in Section \ref{sec:conclusion}.

\subsubsection*{Notations}
In this paper we use the following notations: Scalars are denoted by lowercase or uppercase letters with no special typesetting ($x$), vectors by boldface lowercase letters ($\mathbf x$), sets by blackboard-bold uppercase letter ($\mathbb X$), and matrices by boldface uppercase letters ($\mathbf X$). The element in the $i$-th row and $j$-th column of a matrix $\mathbf A$ is denoted $\mathbf A_{i,j}$. Sequences are indicated by adding subscripts denoting their ranges to the typesetting that matches their elements (e.g., $\mathbf x_{1:k}$ is a sequence of vectors, $\mathbb X_{1:j}$ of sets). The number of elements in a set $\mathbb X$ is denoted $|\mathbb X|$, and we further define $\mathbb N_a\triangleq\{i ~:~ i\leq a~,~ i\in\mathbb N\}$.

\section{MULTI-OBJECT MODELS AND PROBLEM FORMULATION}
\label{sec:preliminaries}
\subsection{Multi-Object Models}
%Throughout this paper we use the standard multi-object dynamic models \cite[Chap.5]{mahler2014}. 
The state vector for an object at time-step $t$ is denoted $\mathbf x_t\in\mathbb R^{d_x}$, and its trajectory is represented by a tuple $\boldsymbol\tau=(t_s, \mathbf x_{t_s:t_s+l})$, where $t_s$ is the initial time in which the object entered the FOV, and $l$ is the number of time-steps it was present for. We use superscripts to specify a certain object $i$, such as $\mathbf x_t^i$ for its state at time $t$. The set of trajectories of all objects inside the FOV at any time-step between $0$ and $T$ is denoted $\mathbb T_T$.
We use the standard multi-object dynamic models \cite[Chap.5]{mahler2014}. 
The birth model is a Poisson point process (PPP) with state-dependent intensity function $\lambda_b(\mathbf x)$, and object death is modeled as independent and identically distributed (i.i.d.) Markovian processes with survival probability $p_s(\mathbf x)$. 
Object motion is also modeled as i.i.d. Markovian, where the single-object transition model is denoted $f(\mathbf x_{t+1}|\mathbf x_t)$. 

We also use the standard measurement models, with the point-object assumption \cite[Chap.5]{mahler2014}. Measurements originating from objects are generated independently for each object, where each object can generate at most one measurement  per time-step with probability of detection $p_d(\mathbf x)$, and each measurement $\mathbf z_t\in\mathbb R^{d_z}$ originates from at most one object. We denote the single-object measurement likelihood as $g(\mathbf z_t|\mathbf x_t)$, and model clutter as a PPP with constant intensity function $\lambda_c(\cdot)$ over the field of view (FOV). The set of all measurements generated at time-step $t$ is denoted $\mathbb Z_t$.

\subsection{Problem Formulation}
\label{sec:problem_formulation}
We focus on the problem of multi-object smoothing for a window of $T$ time-steps. We are given a tuple of measurement sets $(\mathbb Z_1, \cdots, \mathbb Z_T)$ with arbitrary length $T$, and are tasked with estimating the set of object trajectories $\mathbb T_T$. As an example, the tuple of measurement sets could be generated from a radar or LIDAR sensor mounted on top of a vehicle, and $\mathbb T_T$ represents the trajectories of, e.g., pedestrians, cars, and cyclists.

To train a DL solution, we first approximate the multi-object posterior density of $\mathbb T_T$ given $(\mathbb Z_1, \cdots, \mathbb Z_T)$ as a $B$-component multi-Bernoulli RFS density\footnote
{
  A $B$-component multi-Bernoulli RFS density is the disjoint union of $B$ Bernoulli RFS components, each described by an existence probability and a state density function \cite{mahler2014}.
}, where each Bernoulli component $i$ has an existence probability $\bar{p}^i$ and corresponds to one (potential) trajectory. The trajectory for Bernoulli component $i$ is parameterized as a tuple of the form $(\hat{\mathbf x}_{1:T}^i, p_{1:T}^i)$, where $\hat{\mathbf x}_{1:T}^i\in\mathbb R^{d_x\times T}$ is a state sequence of length $T$, and $p_{1:T}^i\in[0,1]^T$ are per-time-step existence probabilities. 

Then, we create the set $\mathbb Z_\Omega$, containing the measurements from all time-steps, together with their times-of-arrival
\begin{equation}
    \mathbb Z_\Omega = \bigcup_{t=1}^T\big\{(\mathbf z, t) : \mathbf z\in\mathbb Z_t\big\}~, \label{eq:TOA}
\end{equation}
and learn models that take as input $\mathbb Z_\Omega$ and output the parameters $\{\hat{\mathbf x}_{1:T}^i, p_{1:T}^i, \bar{p}^i\}_{i=1}^B$ describing the multi-Bernoulli density. In effect, this casts the multi-object smoothing task as a set-to-set prediction task. We also note that the use of existence probabilities $\bar{p}$ allows the models to predict a variable number of trajectories (no larger than $B$), since excess trajectories can have their corresponding $\bar{p}$ set to zero. At the same time, the per-time-step existence probabilities $p_{1:T}$ allow for variable-length trajectories since excess time-steps can have their corresponding $p_t$ set to zero. For example, if $p_t^i=0$ only for $t>t_\text{end}$, then trajectory $i$ is considered to end at time-step $t_\text{end}$, and is only $t_\text{end}$ time-steps long (instead of $T$). In this case, $\hat{\mathbf x}_{t_\text{end}+1:T}$ can be arbitrary vectors not considered part of the trajectory. 

%By forming the set $\mathbb Z_\Omega$ using $(\mathbb Z_1, \cdots, \mathbb Z_T)$, and estimating $\mathbb T_T$ via predicting the parameters of the $B$-component multi-Bernoulli density described above, we cast multi-object smoothing as a set-to-set prediction task: mapping the set $\mathbb Z_\Omega$ to the set of predicted components $\left\{ \hat{\mathbf x}_{1:T}^i, p_{1:T}^i, \bar{p}^i \right\}_{i=1}^B$.

\section{BACKGROUND ON TRANSFORMERS}
\label{sec:background}
This section reviews the transformer encoder architecture used in both the data association and the smoothing modules of D3AS to process sequences of measurements. The transformer model is a popular and successful neural network architecture specifically tailored for sequence-to-sequence learning tasks. %Originally proposed in \cite{transformer_paper}, this model has been used extensively in many different contexts, helping practitioners achieve impressive results in numerous challenging tasks, including translation \cite{transformers_for_translation}, text-to-speech conversion \cite{transformers_in_tts}, sentiment analysis \cite{transformers_in_sentiment_analysis}, protein structure prediction \cite{alphafold}, super-resolution and image generation \cite{image_transformer}, and object detection \cite{DETR, deformable-DETR}, to cite a few.
Originally proposed in \cite{transformer_paper}, its structure is comprised of an encoder and a decoder, but in this paper only the encoder will be reviewed, as this is the only part used in D3AS. The transformer encoder is in charge of processing an input sequence $\mathbf a_{1:n}$, into a new representation $\mathbf e_{1:n}$, referred to as the embeddings of the input sequence, where both $\mathbf a_i,\mathbf e_i\in\mathbb R^d$ for $i\in\mathbb N_n$. After training, the encoder generates embeddings such that each $\mathbf e_i$ encodes the value of the corresponding $\mathbf a_i$ and its relationship to all other elements of the input sequence. Once the embeddings have been computed by the encoder, they can be used by downstream modules that require a global understanding of all elements $\mathbf a_{1:n}$ and their relationships. 

\vspace{-0.1cm}
\subsection{Multihead Self-attention}
The main building block behind the power of transformer architectures is the self-attention layer. This layer processes a sequence $\mathbf a_{1:n}$ into a sequence $\mathbf b_{1:n}$ ($\mathbf a_i, \mathbf b_i\in\mathbb R^d$ for $i\in \mathbb N_n$), and is stacked multiple times inside a transformer encoder. 
%It does so in a manner such that the computation of each $\mathbf b_i$ directly depends on all the input sequence $\mathbf a_{1:n}$, allowing for efficient learning of long-term dependencies in the input. 
The processing of a self-attention layer starts with the computation of three different linear combinations of the input, namely
\begin{equation}
    \label{eq:self-attention_qkv}
    \mathbf{Q} = \mathbf{W}_Q \mathbf{A},~ \mathbf{K} = \mathbf{W}_K\mathbf{A},~ \mathbf{V}=\mathbf{W}_V\mathbf{A},
\end{equation}
where $\mathbf{A} =\begin{bmatrix}\mathbf a_1, \cdots, \mathbf a_n\end{bmatrix} \in \mathbb R^{d\times n}$, and the matrices $\mathbf Q, \mathbf K, \mathbf V$ are referred to as queries, keys, and values, respectively. The matrices $\mathbf{W}_Q, \mathbf{W}_K, \mathbf{W}_V \in \mathbb R^{d\times d}$ are the learnable parameters of the self-attention layer. 
Then, the output is computed as
\begin{equation}
    \mathbf{B}=\mathbf{V}~  \text{Softmax-c}\left(\frac{\mathbf{K}^\top\mathbf{Q}}{\sqrt{d}}\right)~,
\end{equation}
where $\mathbf{B} = \begin{bmatrix}\mathbf b_1, \cdots, \mathbf b_n\end{bmatrix}\in \mathbb R^{d\times n}$ and Softmax-c is the column-wise application of the Softmax function to a matrix, defined as
\begin{align*}
    &[\text{Softmax-c}(\mathbf M)]_{i,j} = \frac{e^{m_{i,j}}}{\sum_{k=1}^{d} e^{m_{k,j}}};\quad i, j\in\mathbb N_n
\end{align*} 
for a matrix $\mathbf{M}\in \mathbb R^{n\times n}$, where $c_{i,j}$ is the element of $\mathbf M$ on row $i$, column $j$.

%The structure behind the computation of $\mathbf b_{1:n}$ reveals why this is such an important part of the transformer. Each of the $\mathbf b_i$ elements directly depends on all inner products of the type $\mathbf a_i^\top \mathbf W\mathbf a_j$, for $j\in\mathbb N_n$, with a learnable $\mathbf W$. Therefore, in contrast to recurrent neural networks, the gradient of any $\mathbf a_i$ w.r.t. to any $\mathbf a_j$ never vanishes, regardless of the length $n$ of the input sequence. This allows the layer to learn about long-term relationships between all the elements of the input sequence, and the more self-attention layers that are stacked (interleaved with nonlinearities), the more complicated relationships the model will be able to learn. 

%We see this property as very important for successful multi-object smoothing, both in the data association step (as a global understanding of the measurements is required for correct associations) and the smoothing step (as the smoothing must take into account information available from all time-steps when estimating the state for any given time).

Additionally, most transformer-based models (including D3AS) use several self-attention layers in parallel and combine the results, referring to the combined computation as a multihead self-attention layer. Concretely, $\mathbf A$ is fed to $n_h$ different self-attention layers with separate learnable parameters, generating $n_h$ different outputs $\mathbf{B}_1, \cdots, \mathbf{B}_{n_h}$.%, all $\in\mathbb R^{d\times n}$.  
The final output $\mathbf B$ is then computed by vertically stacking the results and applying a linear transformation to reduce the dimensionality back to $\mathbb R^{d\times n}$:
\begin{align}
    \mathbf{B} &= \mathbf{W}^0\begin{bmatrix}
        \mathbf{B}_1 \\
        \vdots       \\
        \mathbf{B}_{n_h}\end{bmatrix}~,
\end{align}
where $\mathbf{W}^0\in\mathbb R^{d\times dn_n}$ is a learnable parameter of the multi-head self-attention layer.%, and $\mathbf{B} =\begin{bmatrix}\mathbf b_1, \cdots, \mathbf b_n\end{bmatrix}$.

\subsection{Transformer Encoder}
\label{subsec:transformer_encoder}
\begin{figure}
    \centering

\tikzset{every picture/.style={line width=0.75pt}} %set default line width to 0.75pt        

\begin{tikzpicture}[x=0.75pt,y=0.75pt,yscale=-1,xscale=1]
%uncomment if require: \path (0,600); %set diagram left start at 0, and has height of 600

%Rounded Rect [id:dp48789491851029365] 
\draw  [dash pattern={on 4.5pt off 4.5pt}] (172.6,370.4) .. controls (172.6,355.11) and (184.99,342.72) .. (200.28,342.72) -- (283.32,342.72) .. controls (298.61,342.72) and (311,355.11) .. (311,370.4) -- (311,535.38) .. controls (311,550.66) and (298.61,563.06) .. (283.32,563.06) -- (200.28,563.06) .. controls (184.99,563.06) and (172.6,550.66) .. (172.6,535.38) -- cycle ;
%Flowchart: Or [id:dp373728014203224] 
\draw  [fill={rgb, 255:red, 184; green, 184; blue, 184 }  ,fill opacity=1 ] (239.67,540.83) .. controls (239.67,536.88) and (242.88,533.67) .. (246.83,533.67) .. controls (250.79,533.67) and (254,536.88) .. (254,540.83) .. controls (254,544.79) and (250.79,548) .. (246.83,548) .. controls (242.88,548) and (239.67,544.79) .. (239.67,540.83) -- cycle ; \draw   (239.67,540.83) -- (254,540.83) ; \draw   (246.83,533.67) -- (246.83,548) ;
%Rounded Rect [id:dp10841663659181955] 
\draw  [fill={rgb, 255:red, 249; green, 222; blue, 146 }  ,fill opacity=1 ] (9,382.72) .. controls (9,368.36) and (20.64,356.72) .. (35,356.72) -- (113,356.72) .. controls (127.36,356.72) and (139,368.36) .. (139,382.72) -- (139,526.33) .. controls (139,540.69) and (127.36,552.33) .. (113,552.33) -- (35,552.33) .. controls (20.64,552.33) and (9,540.69) .. (9,526.33) -- cycle ;
%Straight Lines [id:da1942843945735495] 
\draw  [dash pattern={on 4.5pt off 4.5pt}]  (188.78,560.06) -- (130,531.33) ;
%Straight Lines [id:da8085005799217655] 
\draw  [dash pattern={on 4.5pt off 4.5pt}]  (173.44,354.06) -- (130,513.89) ;
%Straight Lines [id:da5969749876586408] 
\draw    (246.83,572.33) -- (246.83,551) ;
\draw [shift={(246.83,548)}, rotate = 90] [fill={rgb, 255:red, 0; green, 0; blue, 0 }  ][line width=0.08]  [draw opacity=0] (6.25,-3) -- (0,0) -- (6.25,3) -- cycle    ;
%Straight Lines [id:da8817697280243637] 
\draw    (271.67,540.83) -- (257,540.83) ;
\draw [shift={(254,540.83)}, rotate = 360] [fill={rgb, 255:red, 0; green, 0; blue, 0 }  ][line width=0.08]  [draw opacity=0] (6.25,-3) -- (0,0) -- (6.25,3) -- cycle    ;
%Straight Lines [id:da972895076460385] 
\draw    (246.83,533.67) -- (246.83,527.39) ;
\draw [shift={(246.83,524.39)}, rotate = 90] [fill={rgb, 255:red, 0; green, 0; blue, 0 }  ][line width=0.08]  [draw opacity=0] (6.25,-3) -- (0,0) -- (6.25,3) -- cycle    ;
%Straight Lines [id:da6391901090259282] 
\draw    (239.67,540.83) -- (184.67,540.83) -- (184.67,441) -- (197,440.73) ;
\draw [shift={(200,440.67)}, rotate = 178.75] [fill={rgb, 255:red, 0; green, 0; blue, 0 }  ][line width=0.08]  [draw opacity=0] (6.25,-3) -- (0,0) -- (6.25,3) -- cycle    ;
%Straight Lines [id:da7753287345377466] 
\draw    (247,476.56) -- (247,463.67) ;
\draw [shift={(247,460.67)}, rotate = 90] [fill={rgb, 255:red, 0; green, 0; blue, 0 }  ][line width=0.08]  [draw opacity=0] (6.25,-3) -- (0,0) -- (6.25,3) -- cycle    ;
%Straight Lines [id:da25810658210075266] 
\draw    (247.5,433.11) -- (247.5,420.33) ;
\draw [shift={(247.5,417.33)}, rotate = 90] [fill={rgb, 255:red, 0; green, 0; blue, 0 }  ][line width=0.08]  [draw opacity=0] (6.25,-3) -- (0,0) -- (6.25,3) -- cycle    ;
%Straight Lines [id:da5881806074608125] 
\draw    (246.83,390.44) -- (246.83,380.72) ;
\draw [shift={(246.83,377.72)}, rotate = 90] [fill={rgb, 255:red, 0; green, 0; blue, 0 }  ][line width=0.08]  [draw opacity=0] (6.25,-3) -- (0,0) -- (6.25,3) -- cycle    ;
%Straight Lines [id:da7259710103386805] 
\draw    (247,426.83) -- (186,426.83) -- (185,361) -- (196.67,361) ;
\draw [shift={(199.67,361)}, rotate = 180] [fill={rgb, 255:red, 0; green, 0; blue, 0 }  ][line width=0.08]  [draw opacity=0] (6.25,-3) -- (0,0) -- (6.25,3) -- cycle    ;
%Straight Lines [id:da4628805880381066] 
\draw    (246.5,349.11) -- (246.5,335.06) ;
\draw [shift={(246.5,332.06)}, rotate = 90] [fill={rgb, 255:red, 0; green, 0; blue, 0 }  ][line width=0.08]  [draw opacity=0] (6.25,-3) -- (0,0) -- (6.25,3) -- cycle    ;
%Straight Lines [id:da4897940705167576] 
\draw    (75.17,560.72) -- (75.17,545.33) ;
\draw [shift={(75.17,542.33)}, rotate = 90] [fill={rgb, 255:red, 0; green, 0; blue, 0 }  ][line width=0.08]  [draw opacity=0] (6.25,-3) -- (0,0) -- (6.25,3) -- cycle    ;
%Straight Lines [id:da6877324814872772] 
\draw    (75.5,512.22) -- (75.5,490.39) ;
\draw [shift={(75.5,487.39)}, rotate = 90] [fill={rgb, 255:red, 0; green, 0; blue, 0 }  ][line width=0.08]  [draw opacity=0] (6.25,-3) -- (0,0) -- (6.25,3) -- cycle    ;
%Straight Lines [id:da5333922376781719] 
\draw    (73.5,384.56) -- (73.5,341.33) ;
\draw [shift={(73.5,338.33)}, rotate = 90] [fill={rgb, 255:red, 0; green, 0; blue, 0 }  ][line width=0.08]  [draw opacity=0] (6.25,-3) -- (0,0) -- (6.25,3) -- cycle    ;
%Rounded Rect [id:dp8296198087244062] 
\draw  [fill={rgb, 255:red, 252; green, 189; blue, 12 }  ,fill opacity=1 ] (20,390.13) .. controls (20,386.93) and (22.6,384.33) .. (25.8,384.33) -- (124.2,384.33) .. controls (127.4,384.33) and (130,386.93) .. (130,390.13) -- (130,407.53) .. controls (130,410.74) and (127.4,413.33) .. (124.2,413.33) -- (25.8,413.33) .. controls (22.6,413.33) and (20,410.74) .. (20,407.53) -- cycle ;
%Rounded Rect [id:dp6709897881409363] 
\draw  [fill={rgb, 255:red, 252; green, 189; blue, 12 }  ,fill opacity=1 ] (20,464.13) .. controls (20,460.93) and (22.6,458.33) .. (25.8,458.33) -- (124.2,458.33) .. controls (127.4,458.33) and (130,460.93) .. (130,464.13) -- (130,481.53) .. controls (130,484.74) and (127.4,487.33) .. (124.2,487.33) -- (25.8,487.33) .. controls (22.6,487.33) and (20,484.74) .. (20,481.53) -- cycle ;
%Rounded Rect [id:dp2587798348442769] 
\draw  [fill={rgb, 255:red, 252; green, 189; blue, 12 }  ,fill opacity=1 ] (20,517.89) .. controls (20,514.69) and (22.6,512.09) .. (25.8,512.09) -- (124.2,512.09) .. controls (127.4,512.09) and (130,514.69) .. (130,517.89) -- (130,535.29) .. controls (130,538.49) and (127.4,541.09) .. (124.2,541.09) -- (25.8,541.09) .. controls (22.6,541.09) and (20,538.49) .. (20,535.29) -- cycle ;
%Rounded Rect [id:dp16615570859437345] 
\draw  [fill={rgb, 255:red, 255; green, 167; blue, 83 }  ,fill opacity=1 ] (197,484.42) .. controls (197,478.97) and (201.42,474.56) .. (206.87,474.56) -- (287.73,474.56) .. controls (293.18,474.56) and (297.6,478.97) .. (297.6,484.42) -- (297.6,514.02) .. controls (297.6,519.47) and (293.18,523.89) .. (287.73,523.89) -- (206.87,523.89) .. controls (201.42,523.89) and (197,519.47) .. (197,514.02) -- cycle ;
%Rounded Rect [id:dp35547639165674316] 
\draw  [fill={rgb, 255:red, 191; green, 183; blue, 255 }  ,fill opacity=1 ] (201.2,439.64) .. controls (201.2,436.59) and (203.68,434.11) .. (206.73,434.11) -- (287.87,434.11) .. controls (290.92,434.11) and (293.4,436.59) .. (293.4,439.64) -- (293.4,456.24) .. controls (293.4,459.3) and (290.92,461.78) .. (287.87,461.78) -- (206.73,461.78) .. controls (203.68,461.78) and (201.2,459.3) .. (201.2,456.24) -- cycle ;
%Rounded Rect [id:dp945090221009597] 
\draw  [fill={rgb, 255:red, 12; green, 252; blue, 234 }  ,fill opacity=1 ] (224,395.51) .. controls (224,392.71) and (226.27,390.44) .. (229.07,390.44) -- (266.33,390.44) .. controls (269.13,390.44) and (271.4,392.71) .. (271.4,395.51) -- (271.4,410.71) .. controls (271.4,413.51) and (269.13,415.78) .. (266.33,415.78) -- (229.07,415.78) .. controls (226.27,415.78) and (224,413.51) .. (224,410.71) -- cycle ;
%Rounded Rect [id:dp9692785865670082] 
\draw  [fill={rgb, 255:red, 191; green, 183; blue, 255 }  ,fill opacity=1 ] (201,354.84) .. controls (201,351.79) and (203.48,349.31) .. (206.53,349.31) -- (287.67,349.31) .. controls (290.72,349.31) and (293.2,351.79) .. (293.2,354.84) -- (293.2,371.44) .. controls (293.2,374.5) and (290.72,376.98) .. (287.67,376.98) -- (206.53,376.98) .. controls (203.48,376.98) and (201,374.5) .. (201,371.44) -- cycle ;
%Straight Lines [id:da7845799656122964] 
\draw    (75.5,459.22) -- (75.5,437.39) ;
\draw [shift={(75.5,434.39)}, rotate = 90] [fill={rgb, 255:red, 0; green, 0; blue, 0 }  ][line width=0.08]  [draw opacity=0] (6.25,-3) -- (0,0) -- (6.25,3) -- cycle    ;

% Text Node
\draw (62.33,560) node [anchor=north west][inner sep=0.75pt]  [font=\normalsize]  {$\mathbf{a}_{1:n}$};
% Text Node
\draw (204,483) node [anchor=north west][inner sep=0.75pt]   [align=left] {\begin{minipage}[lt]{61.12pt}\setlength\topsep{0pt}
\begin{center}
Multihead \\self-attention
\end{center}

\end{minipage}};
% Text Node
\draw (275,537) node [anchor=north west][inner sep=0.75pt]  [font=\small]  {$\mathbf{q}_{1:n}$};
% Text Node
\draw (215,440) node [anchor=north west][inner sep=0.75pt]   [align=left] {LayerNorm};
% Text Node
\draw (233.6,396) node [anchor=north west][inner sep=0.75pt]   [align=left] {FFN};
% Text Node
\draw (215,356) node [anchor=north west][inner sep=0.75pt]   [align=left] {LayerNorm};
% Text Node
\draw (72,409) node [anchor=north west][inner sep=0.75pt]    {$\vdots $};
% Text Node
\draw (32,391.83) node [anchor=north west][inner sep=0.75pt]  [font=\small] [align=left] {Encoder block N};
% Text Node
\draw (33,466) node [anchor=north west][inner sep=0.75pt]  [font=\small] [align=left] {Encoder block 2};
% Text Node
\draw (43,489.73) node [anchor=north west][inner sep=0.75pt]  [font=\small]  {$\mathbf{a}_{1:n}^{( 1)}$};
% Text Node
\draw (33,520) node [anchor=north west][inner sep=0.75pt]  [font=\small] [align=left] {Encoder block 1};
% Text Node
\draw (41.67,436.4) node [anchor=north west][inner sep=0.75pt]  [font=\small]  {$\mathbf{a}_{1:n}^{( 2)}$};
% Text Node
\draw (41,361.07) node [anchor=north west][inner sep=0.75pt]  [font=\small]  {$\mathbf{a}_{1:n}^{( N)}$};
% Text Node
\draw (61,326) node [anchor=north west][inner sep=0.75pt]  [font=\small]  {$\mathbf{e}_{1:n}$};

\end{tikzpicture}

    \caption{The structure of the transformer encoder proposed in \cite{DETR}, where $N$ encoder blocks are connected in series. The components of each encoder block are shown on the right.}
    \label{fig:encoder}
    \vspace{-10pt}
\end{figure}
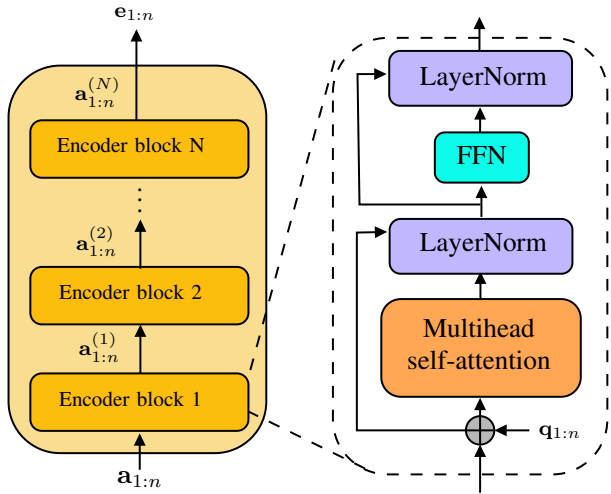
Stacking several layers of multiheaded self-attention allows models to learn complicated long-term dependencies in input sequence. The transformer encoder does exactly this, interleaving multiheaded self-attention layers with other types of nonlinearities. In this paper, we use a slightly modified version of the encoder, proposed in \cite{DETR}, as shown in Fig.\,\ref{fig:encoder}. The overall encoder is built by stacking $N$ ``encoder blocks'', where the output from one block is fed to the next. The computation for encoder block $l\in\mathbb N_N$ can be described as 
\begin{align}
    \label{eq:encoder_eq1}
    &\tilde{\mathbf a}_{1:n}^{(l-1)} = \mathbf a_{1:n}^{(l-1)} + \mathbf q_{1:n}
    \\
    \label{eq:encoder_eq2}
    &\mathbf b^{(l)}_{1:n} = \mathrm{MultiHeadAttention}(\tilde{\mathbf a}_{1:n}^{(l-1)})
    \\
    \label{eq:encoder_eq3}
    &\tilde{\mathbf{b}}^{(l)}_{1:n} = \mathrm{LayerNorm}(\tilde{\mathbf a}_{1:n}^{(l-1)} + \mathbf b^{(l)}_{1:n})
    \\
    \label{eq:encoder_eq4}
    &\mathbf a_{1:n}^{(l)} = \mathrm{LayerNorm}(
    \tilde{\mathbf{b}}^{(l)}_{1:n} + \text{FFN}(\tilde{\mathbf{b}}^{(l)}_{1:n}))~,
\end{align}
where MultiHeadAttention is a multi-head self-attention layer, FFN is a fully-connected feedforward neural network applied to each element of the input sequence separately, and LayerNorm is a Layer Normalization layer \cite{layer_normalization}. 
We use the notation $\mathbf a_{1:n}^{(l)}$ to denote the input sequence after being processed by $l$ encoder blocks. For instance, $\mathbf a_{1:n}^{(0)}$ is the original input sequence $\mathbf a_{1:n}$, and $\mathbf a_{1:n}^{(N)}$ is the output of the encoder module, also denoted $\mathbf e_{1:n}$. 
Additionally, at the start of every encoder block, the sequence $\mathbf z_{1:n}^{(l)}$ is added with a vector $\mathbf q_{1:n}$, referred to as the positional encoding, which can be either fixed (usually with sinusoidal components) \cite{transformer_paper} or learnable \cite{DETR}. Without this addition, the encoder would become permutation-equivariant\footnote{A function $f$ is equivariant to a transformation $g$ iff $f(g(x))=g(f(x)).$}, which is undesirable in many contexts that rely on the relative position of the elements $\mathbf a_i$ in $\mathbf a_{1:n}$ to convey important information.

\section{METHOD}
\label{sec:method}
This section describes D3AS, the proposed method for performing multi-object smoothing in the model-based context. It is divided into two modules: the Deep Data Associator (DDA) and the Deep Smoother (DS).
\subsection{Overview of DDA and DS modules}
\begin{figure*}
    \centering

    \tikzset{every picture/.style={line width=0.75pt}} %set default line width to 0.75pt        
    \resizebox{0.9\textwidth}{!}
    {
            
        \begin{tikzpicture}[x=0.75pt,y=0.75pt,yscale=-1,xscale=1]
        %uncomment if require: \path (0,391); %set diagram left start at 0, and has height of 391
        
        %Shape: Rectangle [id:dp47047713534048374] 
        \draw  [fill={rgb, 255:red, 255; green, 255; blue, 255 }  ,fill opacity=1 ][dash pattern={on 0.84pt off 2.51pt}] (380.25,130.03) -- (490.27,130.03) -- (490.27,181.73) -- (380.25,181.73) -- cycle ;
        %Shape: Rectangle [id:dp9610732963988236] 
        \draw  [fill={rgb, 255:red, 255; green, 255; blue, 255 }  ,fill opacity=1 ][dash pattern={on 0.84pt off 2.51pt}] (379.65,70.43) -- (489.67,70.43) -- (489.67,122.13) -- (379.65,122.13) -- cycle ;
        %Shape: Rectangle [id:dp7929767723377508] 
        \draw  [fill={rgb, 255:red, 255; green, 255; blue, 255 }  ,fill opacity=1 ][dash pattern={on 0.84pt off 2.51pt}] (381.65,190.63) -- (491.67,190.63) -- (491.67,242.33) -- (381.65,242.33) -- cycle ;
        %Shape: Rectangle [id:dp7530657215149872] 
        \draw  [fill={rgb, 255:red, 235; green, 34; blue, 254 }  ,fill opacity=0.7 ] (387.03,217.48) -- (403.39,217.48) -- (403.39,233.84) -- (387.03,233.84) -- cycle ;
        %Shape: Rectangle [id:dp9189103118533857] 
        \draw  [fill={rgb, 255:red, 0; green, 0; blue, 0 }  ,fill opacity=0.7 ] (407.85,217.48) -- (424.21,217.48) -- (424.21,233.84) -- (407.85,233.84) -- cycle ;
        %Shape: Rectangle [id:dp06394234098079621] 
        \draw  [fill={rgb, 255:red, 0; green, 0; blue, 0 }  ,fill opacity=0.7 ] (428.67,217.48) -- (445.03,217.48) -- (445.03,233.84) -- (428.67,233.84) -- cycle ;
        %Shape: Rectangle [id:dp6647506138803554] 
        \draw  [fill={rgb, 255:red, 235; green, 34; blue, 254 }  ,fill opacity=0.7 ] (470.31,217.48) -- (486.67,217.48) -- (486.67,233.84) -- (470.31,233.84) -- cycle ;
        %Shape: Rectangle [id:dp3058059160730988] 
        \draw  [fill={rgb, 255:red, 0; green, 0; blue, 0 }  ,fill opacity=0.7 ] (449.49,217.65) -- (465.85,217.65) -- (465.85,234.01) -- (449.49,234.01) -- cycle ;
        %Shape: Rectangle [id:dp2444757967361617] 
        \draw  [fill={rgb, 255:red, 235; green, 34; blue, 254 }  ,fill opacity=0.7 ] (383.43,95.98) -- (400.37,95.98) -- (400.37,112.93) -- (383.43,112.93) -- cycle ;
        %Shape: Rectangle [id:dp1491175068656767] 
        \draw  [fill={rgb, 255:red, 235; green, 34; blue, 254 }  ,fill opacity=0.7 ] (405.89,95.98) -- (422.83,95.98) -- (422.83,112.93) -- (405.89,112.93) -- cycle ;
        %Shape: Rectangle [id:dp3750157823528675] 
        \draw  [fill={rgb, 255:red, 235; green, 34; blue, 254 }  ,fill opacity=0.7 ] (427.46,95.98) -- (444.4,95.98) -- (444.4,112.93) -- (427.46,112.93) -- cycle ;
        %Shape: Rectangle [id:dp07279264001195695] 
        \draw  [fill={rgb, 255:red, 0; green, 0; blue, 0 }  ,fill opacity=0.7 ] (470.58,95.98) -- (487.53,95.98) -- (487.53,112.93) -- (470.58,112.93) -- cycle ;
        %Shape: Rectangle [id:dp2080470658277902] 
        \draw  [fill={rgb, 255:red, 235; green, 34; blue, 254 }  ,fill opacity=0.7 ] (449.02,95.98) -- (465.96,95.98) -- (465.96,112.93) -- (449.02,112.93) -- cycle ;
        %Shape: Rectangle [id:dp511304151646752] 
        \draw  [fill={rgb, 255:red, 0; green, 0; blue, 0 }  ,fill opacity=0.4 ] (385.37,159.34) -- (401.68,159.34) -- (401.68,175.65) -- (385.37,175.65) -- cycle ;
        %Shape: Rectangle [id:dp3470253876653928] 
        \draw  [fill={rgb, 255:red, 0; green, 0; blue, 0 }  ,fill opacity=0.7 ] (407,159.34) -- (423.31,159.34) -- (423.31,175.65) -- (407,175.65) -- cycle ;
        %Shape: Rectangle [id:dp1327655295880643] 
        \draw  [fill={rgb, 255:red, 0; green, 0; blue, 0 }  ,fill opacity=0.7 ] (427.76,159.34) -- (444.08,159.34) -- (444.08,175.65) -- (427.76,175.65) -- cycle ;
        %Shape: Rectangle [id:dp645644409761212] 
        \draw  [fill={rgb, 255:red, 0; green, 0; blue, 0 }  ,fill opacity=0.7 ] (469.29,159.34) -- (485.61,159.34) -- (485.61,175.65) -- (469.29,175.65) -- cycle ;
        %Shape: Rectangle [id:dp15769561236612173] 
        \draw  [fill={rgb, 255:red, 235; green, 34; blue, 254 }  ,fill opacity=0.7 ] (448.53,159.34) -- (464.84,159.34) -- (464.84,175.65) -- (448.53,175.65) -- cycle ;
        %Shape: Grid [id:dp5070307861255929] 
        \draw  [draw opacity=0] (16,122.76) -- (137.55,122.76) -- (137.55,220) -- (16,220) -- cycle ; \draw  [color={rgb, 255:red, 0; green, 0; blue, 0 }  ,draw opacity=0.1 ] (40.31,122.76) -- (40.31,220)(64.62,122.76) -- (64.62,220)(88.93,122.76) -- (88.93,220)(113.24,122.76) -- (113.24,220) ; \draw  [color={rgb, 255:red, 0; green, 0; blue, 0 }  ,draw opacity=0.1 ] (16,147.07) -- (137.55,147.07)(16,171.38) -- (137.55,171.38)(16,195.69) -- (137.55,195.69) ; \draw  [color={rgb, 255:red, 0; green, 0; blue, 0 }  ,draw opacity=0.1 ] (16,122.76) -- (137.55,122.76) -- (137.55,220) -- (16,220) -- cycle ;
        %Shape: Rectangle [id:dp30513599396117486] 
        \draw  [fill={rgb, 255:red, 0; green, 0; blue, 0 }  ,fill opacity=0 ] (238.83,93.89) -- (253.62,93.89) -- (253.62,110.73) -- (238.83,110.73) -- cycle ;
        %Shape: Rectangle [id:dp13373687438975423] 
        \draw  [fill={rgb, 255:red, 0; green, 0; blue, 0 }  ,fill opacity=0.4 ] (253.62,93.89) -- (268.4,93.89) -- (268.4,110.73) -- (253.62,110.73) -- cycle ;
        %Shape: Rectangle [id:dp881436056231998] 
        \draw  [fill={rgb, 255:red, 0; green, 0; blue, 0 }  ,fill opacity=0.3 ] (268.4,93.89) -- (283.18,93.89) -- (283.18,110.73) -- (268.4,110.73) -- cycle ;
        %Shape: Rectangle [id:dp9926500606951476] 
        \draw  [fill={rgb, 255:red, 0; green, 0; blue, 0 }  ,fill opacity=0 ] (283.18,93.89) -- (297.97,93.89) -- (297.97,110.73) -- (283.18,110.73) -- cycle ;
        %Shape: Rectangle [id:dp9021662442533613] 
        \draw  [fill={rgb, 255:red, 0; green, 0; blue, 0 }  ,fill opacity=0.7 ] (268.4,110.73) -- (283.18,110.73) -- (283.18,127.57) -- (268.4,127.57) -- cycle ;
        %Shape: Rectangle [id:dp26638965161712824] 
        \draw  [fill={rgb, 255:red, 0; green, 0; blue, 0 }  ,fill opacity=0 ] (283.18,110.73) -- (297.97,110.73) -- (297.97,127.57) -- (283.18,127.57) -- cycle ;
        %Shape: Rectangle [id:dp338055247580725] 
        \draw  [fill={rgb, 255:red, 0; green, 0; blue, 0 }  ,fill opacity=0 ] (238.83,110.73) -- (253.62,110.73) -- (253.62,127.57) -- (238.83,127.57) -- cycle ;
        %Shape: Rectangle [id:dp7908024605328283] 
        \draw  [fill={rgb, 255:red, 0; green, 0; blue, 0 }  ,fill opacity=0 ] (238.83,127.57) -- (253.62,127.57) -- (253.62,144.41) -- (238.83,144.41) -- cycle ;
        %Shape: Rectangle [id:dp165913070788118] 
        \draw  [fill={rgb, 255:red, 0; green, 0; blue, 0 }  ,fill opacity=0.7 ] (253.62,127.57) -- (268.4,127.57) -- (268.4,144.41) -- (253.62,144.41) -- cycle ;
        %Shape: Rectangle [id:dp8935995124225034] 
        \draw  [fill={rgb, 255:red, 0; green, 0; blue, 0 }  ,fill opacity=0 ] (268.4,127.57) -- (283.18,127.57) -- (283.18,144.41) -- (268.4,144.41) -- cycle ;
        %Shape: Rectangle [id:dp091683012323029] 
        \draw  [fill={rgb, 255:red, 0; green, 0; blue, 0 }  ,fill opacity=0 ] (283.18,127.57) -- (297.97,127.57) -- (297.97,144.41) -- (283.18,144.41) -- cycle ;
        %Shape: Rectangle [id:dp7046671248373351] 
        \draw  [fill={rgb, 255:red, 0; green, 0; blue, 0 }  ,fill opacity=0.7 ] (253.62,144.41) -- (268.4,144.41) -- (268.4,161.24) -- (253.62,161.24) -- cycle ;
        %Shape: Rectangle [id:dp7027933676573406] 
        \draw  [fill={rgb, 255:red, 0; green, 0; blue, 0 }  ,fill opacity=0 ] (268.4,144.41) -- (283.18,144.41) -- (283.18,161.24) -- (268.4,161.24) -- cycle ;
        %Shape: Rectangle [id:dp4326030505767844] 
        \draw  [fill={rgb, 255:red, 0; green, 0; blue, 0 }  ,fill opacity=0 ] (283.18,144.41) -- (297.97,144.41) -- (297.97,161.24) -- (283.18,161.24) -- cycle ;
        %Shape: Rectangle [id:dp07041571258011348] 
        \draw  [fill={rgb, 255:red, 0; green, 0; blue, 0 }  ,fill opacity=0 ] (238.83,144.41) -- (253.62,144.41) -- (253.62,161.24) -- (238.83,161.24) -- cycle ;
        %Shape: Rectangle [id:dp48679041869697715] 
        \draw  [fill={rgb, 255:red, 0; green, 0; blue, 0 }  ,fill opacity=0 ] (238.83,161.24) -- (253.62,161.24) -- (253.62,178.08) -- (238.83,178.08) -- cycle ;
        %Shape: Rectangle [id:dp1184021759966225] 
        \draw  [fill={rgb, 255:red, 0; green, 0; blue, 0 }  ,fill opacity=0 ] (253.62,161.24) -- (268.4,161.24) -- (268.4,178.08) -- (253.62,178.08) -- cycle ;
        %Shape: Rectangle [id:dp2804941111818504] 
        \draw  [fill={rgb, 255:red, 0; green, 0; blue, 0 }  ,fill opacity=0.7 ] (268.4,161.24) -- (283.18,161.24) -- (283.18,178.08) -- (268.4,178.08) -- cycle ;
        %Shape: Rectangle [id:dp6424887170940359] 
        \draw  [fill={rgb, 255:red, 0; green, 0; blue, 0 }  ,fill opacity=0 ] (283.18,161.24) -- (297.97,161.24) -- (297.97,178.08) -- (283.18,178.08) -- cycle ;
        %Shape: Rectangle [id:dp5274240247406861] 
        \draw  [fill={rgb, 255:red, 0; green, 0; blue, 0 }  ,fill opacity=0 ] (253.62,178.08) -- (268.4,178.08) -- (268.4,194.92) -- (253.62,194.92) -- cycle ;
        %Shape: Rectangle [id:dp2557253101991137] 
        \draw  [fill={rgb, 255:red, 0; green, 0; blue, 0 }  ,fill opacity=0.7 ] (268.4,178.08) -- (283.18,178.08) -- (283.18,194.92) -- (268.4,194.92) -- cycle ;
        %Shape: Rectangle [id:dp801023011790329] 
        \draw  [fill={rgb, 255:red, 0; green, 0; blue, 0 }  ,fill opacity=0 ] (283.18,178.08) -- (297.97,178.08) -- (297.97,194.92) -- (283.18,194.92) -- cycle ;
        %Shape: Rectangle [id:dp4430187926937288] 
        \draw  [fill={rgb, 255:red, 0; green, 0; blue, 0 }  ,fill opacity=0 ] (238.83,178.08) -- (253.62,178.08) -- (253.62,194.92) -- (238.83,194.92) -- cycle ;
        %Shape: Rectangle [id:dp9777281611064785] 
        \draw  [fill={rgb, 255:red, 0; green, 0; blue, 0 }  ,fill opacity=0 ] (238.83,194.92) -- (253.62,194.92) -- (253.62,211.76) -- (238.83,211.76) -- cycle ;
        %Shape: Rectangle [id:dp8691900175302893] 
        \draw  [fill={rgb, 255:red, 0; green, 0; blue, 0 }  ,fill opacity=0.7 ] (253.62,194.92) -- (268.4,194.92) -- (268.4,211.76) -- (253.62,211.76) -- cycle ;
        %Shape: Rectangle [id:dp10889097350072974] 
        \draw  [fill={rgb, 255:red, 0; green, 0; blue, 0 }  ,fill opacity=0 ] (268.4,194.92) -- (283.18,194.92) -- (283.18,211.76) -- (268.4,211.76) -- cycle ;
        %Shape: Rectangle [id:dp30295048677735315] 
        \draw  [fill={rgb, 255:red, 0; green, 0; blue, 0 }  ,fill opacity=0 ] (283.18,194.92) -- (297.97,194.92) -- (297.97,211.76) -- (283.18,211.76) -- cycle ;
        %Shape: Rectangle [id:dp11297789705414418] 
        \draw  [fill={rgb, 255:red, 0; green, 0; blue, 0 }  ,fill opacity=0 ] (253.62,211.76) -- (268.4,211.76) -- (268.4,228.6) -- (253.62,228.6) -- cycle ;
        %Shape: Rectangle [id:dp8641197142804697] 
        \draw  [fill={rgb, 255:red, 0; green, 0; blue, 0 }  ,fill opacity=0 ] (268.4,211.76) -- (283.18,211.76) -- (283.18,228.6) -- (268.4,228.6) -- cycle ;
        %Shape: Rectangle [id:dp35601718489758394] 
        \draw  [fill={rgb, 255:red, 0; green, 0; blue, 0 }  ,fill opacity=0 ] (283.18,211.76) -- (297.97,211.76) -- (297.97,228.6) -- (283.18,228.6) -- cycle ;
        %Shape: Rectangle [id:dp40576003575304] 
        \draw  [fill={rgb, 255:red, 0; green, 0; blue, 0 }  ,fill opacity=0.7 ] (238.83,211.76) -- (253.62,211.76) -- (253.62,228.6) -- (238.83,228.6) -- cycle ;
        %Shape: Rectangle [id:dp19809783170517314] 
        \draw  [fill={rgb, 255:red, 0; green, 0; blue, 0 }  ,fill opacity=0 ] (253.62,110.73) -- (268.4,110.73) -- (268.4,127.57) -- (253.62,127.57) -- cycle ;
        %Straight Lines [id:da5996045824557326] 
        \draw    (297.75,166.75) -- (312,166.75) ;
        \draw [shift={(315,166.75)}, rotate = 180] [fill={rgb, 255:red, 0; green, 0; blue, 0 }  ][line width=0.08]  [draw opacity=0] (8.93,-4.29) -- (0,0) -- (8.93,4.29) -- cycle    ;
        \draw  [color={rgb, 255:red, 0; green, 0; blue, 0 }  ,draw opacity=1 ][line width=0.75]  (86.43,188.06) -- (91.91,182.63)(86.46,182.6) -- (91.88,188.09) ;
        \draw  [color={rgb, 255:red, 0; green, 0; blue, 0 }  ,draw opacity=1 ][line width=0.75]  (110.97,202.61) -- (116.45,197.18)(111,197.15) -- (116.42,202.64) ;
        \draw  [color={rgb, 255:red, 0; green, 0; blue, 0 }  ,draw opacity=1 ][line width=0.75]  (134.95,140.63) -- (140.44,135.2)(134.98,135.17) -- (140.41,140.66) ;
        \draw  [color={rgb, 255:red, 0; green, 0; blue, 0 }  ,draw opacity=1 ][line width=0.75]  (86.57,151.52) -- (92.06,146.09)(86.6,146.06) -- (92.03,151.55) ;
        \draw  [color={rgb, 255:red, 0; green, 0; blue, 0 }  ,draw opacity=1 ][line width=0.75]  (62.45,159) -- (67.93,153.58)(62.48,153.55) -- (67.9,159.03) ;
        \draw  [color={rgb, 255:red, 0; green, 0; blue, 0 }  ,draw opacity=1 ][line width=0.75]  (62.68,177.78) -- (68.16,172.35)(62.71,172.32) -- (68.13,177.81) ;
        %Straight Lines [id:da33208437984679917] 
        \draw    (16,220) -- (16,107) ;
        \draw [shift={(16,104)}, rotate = 90] [fill={rgb, 255:red, 0; green, 0; blue, 0 }  ][line width=0.08]  [draw opacity=0] (8.93,-4.29) -- (0,0) -- (8.93,4.29) -- cycle    ;
        %Straight Lines [id:da33964543070925934] 
        \draw    (16,219.43) -- (150.59,219.43) ;
        \draw [shift={(153.59,219.43)}, rotate = 180] [fill={rgb, 255:red, 0; green, 0; blue, 0 }  ][line width=0.08]  [draw opacity=0] (8.93,-4.29) -- (0,0) -- (8.93,4.29) -- cycle    ;
        %Straight Lines [id:da4278373872862966] 
        \draw    (146,168) -- (164.33,168) ;
        \draw [shift={(167.33,168)}, rotate = 180] [fill={rgb, 255:red, 0; green, 0; blue, 0 }  ][line width=0.08]  [draw opacity=0] (8.93,-4.29) -- (0,0) -- (8.93,4.29) -- cycle    ;
        %Straight Lines [id:da9572783906162372] 
        \draw    (209.67,167) -- (224.5,167) ;
        \draw [shift={(227.5,167)}, rotate = 180] [fill={rgb, 255:red, 0; green, 0; blue, 0 }  ][line width=0.08]  [draw opacity=0] (8.93,-4.29) -- (0,0) -- (8.93,4.29) -- cycle    ;
        %Straight Lines [id:da3564035583457097] 
        \draw    (356.07,219.33) -- (378,219.33) ;
        \draw [shift={(381,219.33)}, rotate = 180] [fill={rgb, 255:red, 0; green, 0; blue, 0 }  ][line width=0.08]  [draw opacity=0] (8.93,-4.29) -- (0,0) -- (8.93,4.29) -- cycle    ;
        \draw  [color={rgb, 255:red, 0; green, 0; blue, 0 }  ,draw opacity=1 ][line width=0.75]  (134.78,173.07) -- (140.27,167.64)(134.82,167.61) -- (140.24,173.1) ;
        \draw  [color={rgb, 255:red, 0; green, 0; blue, 0 }  ,draw opacity=1 ][line width=0.75]  (37.18,187.78) -- (42.66,182.35)(37.21,182.32) -- (42.63,187.81) ;
        %Straight Lines [id:da08695078011818458] 
        \draw    (356.07,227.33) -- (356.07,152.67) -- (378,152.67) ;
        \draw [shift={(381,152.67)}, rotate = 180] [fill={rgb, 255:red, 0; green, 0; blue, 0 }  ][line width=0.08]  [draw opacity=0] (8.93,-4.29) -- (0,0) -- (8.93,4.29) -- cycle    ;
        %Straight Lines [id:da2351335909595571] 
        \draw    (356.07,198.33) -- (356.07,261) -- (566,261) ;
        \draw [shift={(569,261)}, rotate = 180] [fill={rgb, 255:red, 0; green, 0; blue, 0 }  ][line width=0.08]  [draw opacity=0] (8.93,-4.29) -- (0,0) -- (8.93,4.29) -- cycle    ;
        %Straight Lines [id:da44683873057610657] 
        \draw    (489,155.33) -- (505.79,155.33) ;
        \draw [shift={(508.79,155.33)}, rotate = 180] [fill={rgb, 255:red, 0; green, 0; blue, 0 }  ][line width=0.08]  [draw opacity=0] (8.93,-4.29) -- (0,0) -- (8.93,4.29) -- cycle    ;
        %Straight Lines [id:da08583861277353177] 
        \draw    (549,156.33) -- (563.79,156.33) ;
        \draw [shift={(566.79,156.33)}, rotate = 180] [fill={rgb, 255:red, 0; green, 0; blue, 0 }  ][line width=0.08]  [draw opacity=0] (8.93,-4.29) -- (0,0) -- (8.93,4.29) -- cycle    ;
        %Straight Lines [id:da5357447616627296] 
        \draw    (492.6,217.53) -- (507.6,217.53) ;
        \draw [shift={(510.6,217.53)}, rotate = 180] [fill={rgb, 255:red, 0; green, 0; blue, 0 }  ][line width=0.08]  [draw opacity=0] (8.93,-4.29) -- (0,0) -- (8.93,4.29) -- cycle    ;
        %Straight Lines [id:da46926831691824167] 
        \draw    (552.6,218.53) -- (567.6,218.53) ;
        \draw [shift={(570.6,218.53)}, rotate = 180] [fill={rgb, 255:red, 0; green, 0; blue, 0 }  ][line width=0.08]  [draw opacity=0] (8.93,-4.29) -- (0,0) -- (8.93,4.29) -- cycle    ;
        %Straight Lines [id:da7758745225769277] 
        \draw    (490.6,97.13) -- (505.6,97.13) ;
        \draw [shift={(508.6,97.13)}, rotate = 180] [fill={rgb, 255:red, 0; green, 0; blue, 0 }  ][line width=0.08]  [draw opacity=0] (8.93,-4.29) -- (0,0) -- (8.93,4.29) -- cycle    ;
        %Straight Lines [id:da3369518639352078] 
        \draw    (550.6,98.13) -- (565.6,98.13) ;
        \draw [shift={(568.6,98.13)}, rotate = 180] [fill={rgb, 255:red, 0; green, 0; blue, 0 }  ][line width=0.08]  [draw opacity=0] (8.93,-4.29) -- (0,0) -- (8.93,4.29) -- cycle    ;
        %Straight Lines [id:da7371272048581483] 
        \draw    (356.07,170.67) -- (356.07,96) -- (378,96) ;
        \draw [shift={(381,96)}, rotate = 180] [fill={rgb, 255:red, 0; green, 0; blue, 0 }  ][line width=0.08]  [draw opacity=0] (8.93,-4.29) -- (0,0) -- (8.93,4.29) -- cycle    ;
        %Straight Lines [id:da3321649134586351] 
        \draw    (338,167) -- (356,167) ;
        %Rounded Rect [id:dp9442888650999475] 
        \draw  [fill={rgb, 255:red, 255; green, 149; blue, 149 }  ,fill opacity=1 ] (166.78,159.58) .. controls (166.78,156.71) and (169.1,154.39) .. (171.97,154.39) -- (203.59,154.39) .. controls (206.45,154.39) and (208.78,156.71) .. (208.78,159.58) -- (208.78,175.14) .. controls (208.78,178.01) and (206.45,180.33) .. (203.59,180.33) -- (171.97,180.33) .. controls (169.1,180.33) and (166.78,178.01) .. (166.78,175.14) -- cycle ;
        %Rounded Rect [id:dp35989307403956394] 
        \draw  [fill={rgb, 255:red, 201; green, 201; blue, 201 }  ,fill opacity=1 ] (315.33,159.49) .. controls (315.33,157.01) and (317.35,155) .. (319.83,155) -- (333.31,155) .. controls (335.79,155) and (337.8,157.01) .. (337.8,159.49) -- (337.8,175.84) .. controls (337.8,178.32) and (335.79,180.33) .. (333.31,180.33) -- (319.83,180.33) .. controls (317.35,180.33) and (315.33,178.32) .. (315.33,175.84) -- cycle ;
        %Rounded Rect [id:dp3929071262902072] 
        \draw  [fill={rgb, 255:red, 145; green, 245; blue, 101 }  ,fill opacity=1 ] (508.13,87.2) .. controls (508.13,83.67) and (511,80.8) .. (514.53,80.8) -- (544.07,80.8) .. controls (547.6,80.8) and (550.47,83.67) .. (550.47,87.2) -- (550.47,106.4) .. controls (550.47,109.93) and (547.6,112.8) .. (544.07,112.8) -- (514.53,112.8) .. controls (511,112.8) and (508.13,109.93) .. (508.13,106.4) -- cycle ;
        %Rounded Rect [id:dp9144228529689347] 
        \draw  [fill={rgb, 255:red, 145; green, 245; blue, 101 }  ,fill opacity=1 ] (508.13,146.2) .. controls (508.13,142.67) and (511,139.8) .. (514.53,139.8) -- (544.07,139.8) .. controls (547.6,139.8) and (550.47,142.67) .. (550.47,146.2) -- (550.47,165.4) .. controls (550.47,168.93) and (547.6,171.8) .. (544.07,171.8) -- (514.53,171.8) .. controls (511,171.8) and (508.13,168.93) .. (508.13,165.4) -- cycle ;
        %Rounded Rect [id:dp5221548901948596] 
        \draw  [fill={rgb, 255:red, 145; green, 245; blue, 101 }  ,fill opacity=1 ] (509.13,208.53) .. controls (509.13,205) and (512,202.13) .. (515.53,202.13) -- (545.07,202.13) .. controls (548.6,202.13) and (551.47,205) .. (551.47,208.53) -- (551.47,227.73) .. controls (551.47,231.27) and (548.6,234.13) .. (545.07,234.13) -- (515.53,234.13) .. controls (512,234.13) and (509.13,231.27) .. (509.13,227.73) -- cycle ;
        
        % Text Node
        \draw (174,161) node [anchor=north west][inner sep=0.75pt]  [font=\small] [align=left] {DDA};
        % Text Node
        \draw (221.59,234.98) node [anchor=north west][inner sep=0.75pt]  [font=\small] [align=left] {\begin{minipage}[lt]{70.63pt}\setlength\topsep{0pt}
        \begin{center}
        Data association\\matrix
        \end{center}
        
        \end{minipage}};
        % Text Node
        \draw (67,225) node [anchor=north west][inner sep=0.75pt]  [font=\small] [align=left] {\begin{minipage}[lt]{22.79pt}\setlength\topsep{0pt}
        \begin{center}
        Time
        \end{center}
        
        \end{minipage}};
        % Text Node
        \draw (-0.81,183.91) node [anchor=north west][inner sep=0.75pt]  [font=\small,rotate=-270.73] [align=left] {\begin{minipage}[lt]{35.39pt}\setlength\topsep{0pt}
        \begin{center}
        Position
        \end{center}
        
        \end{minipage}};
        % Text Node
        \draw (61.2,159.4) node [anchor=north west][inner sep=0.75pt]  [font=\footnotesize] [align=left] {2};
        % Text Node
        \draw (60.8,140.6) node [anchor=north west][inner sep=0.75pt]  [font=\footnotesize] [align=left] {3};
        % Text Node
        \draw (85.6,132) node [anchor=north west][inner sep=0.75pt]  [font=\footnotesize] [align=left] {4};
        % Text Node
        \draw (85.4,169.8) node [anchor=north west][inner sep=0.75pt]  [font=\footnotesize] [align=left] {5};
        % Text Node
        \draw (109,184.6) node [anchor=north west][inner sep=0.75pt]  [font=\footnotesize] [align=left] {6};
        % Text Node
        \draw (134.2,120.8) node [anchor=north west][inner sep=0.75pt]  [font=\footnotesize] [align=left] {7};
        % Text Node
        \draw (133.8,155.8) node [anchor=north west][inner sep=0.75pt]  [font=\footnotesize] [align=left] {8};
        % Text Node
        \draw (35.8,168.8) node [anchor=north west][inner sep=0.75pt]  [font=\footnotesize] [align=left] {1};
        % Text Node
        \draw (384.69,134.98) node [anchor=north west][inner sep=0.75pt]    {$\mathbf{z}_{1}$};
        % Text Node
        \draw (408.36,134.98) node [anchor=north west][inner sep=0.75pt]    {$\mathbf{z}_{3}$};
        % Text Node
        \draw (428.57,134.98) node [anchor=north west][inner sep=0.75pt]    {$\mathbf{z}_{4}$};
        % Text Node
        \draw (470.95,134.98) node [anchor=north west][inner sep=0.75pt]    {$\mathbf{z}_{7}$};
        % Text Node
        \draw (450.9,132.71) node [anchor=north west][inner sep=0.75pt]    {$\tilde{\mathbf{z}}$};
        % Text Node
        \draw (389.31,192.95) node [anchor=north west][inner sep=0.75pt]    {$\tilde{\mathbf{z}}$};
        % Text Node
        \draw (408.37,196.55) node [anchor=north west][inner sep=0.75pt]    {$\mathbf{z}_{2}$};
        % Text Node
        \draw (428.64,196.55) node [anchor=north west][inner sep=0.75pt]    {$\mathbf{z}_{5}$};
        % Text Node
        \draw (472.32,192.95) node [anchor=north west][inner sep=0.75pt]    {$\tilde{\mathbf{z}}$};
        % Text Node
        \draw (449.8,196.55) node [anchor=north west][inner sep=0.75pt]    {$\mathbf{z}_{6}{}$};
        % Text Node
        \draw (386.01,70.92) node [anchor=north west][inner sep=0.75pt]    {$\tilde{\mathbf{z}}$};
        % Text Node
        \draw (408.79,70.92) node [anchor=north west][inner sep=0.75pt]    {$\tilde{\mathbf{z}}$};
        % Text Node
        \draw (429.78,70.92) node [anchor=north west][inner sep=0.75pt]    {$\tilde{\mathbf{z}}$};
        % Text Node
        \draw (472.63,74.07) node [anchor=north west][inner sep=0.75pt]    {$\mathbf{z}_{8}$};
        % Text Node
        \draw (451.99,70.61) node [anchor=north west][inner sep=0.75pt]    {$\tilde{\mathbf{z}}$};
        % Text Node
        \draw (322,161) node [anchor=north west][inner sep=0.75pt]  [font=\small] [align=left] {P};
        % Text Node
        \draw (571.6,208.4) node [anchor=north west][inner sep=0.75pt]    {$\hat{\mathbf{x}}_{1:T}^{3} ,p_{1:T}^{3} ,\overline{p}^{3}$};
        % Text Node
        \draw (570.6,145.8) node [anchor=north west][inner sep=0.75pt]    {$\hat{\mathbf{x}}_{1:T}^{2} ,p_{1:T}^{2} ,\overline{p}^{2}$};
        % Text Node
        \draw (571.6,87.6) node [anchor=north west][inner sep=0.75pt]    {$\hat{\mathbf{x}}_{1:T}^{1} ,p_{1:T}^{1} ,\overline{p}^{1}$};
        % Text Node
        \draw (521,149) node [anchor=north west][inner sep=0.75pt]  [font=\small] [align=left] {DS};
        % Text Node
        \draw (521,211) node [anchor=north west][inner sep=0.75pt]  [font=\small] [align=left] {DS};
        % Text Node
        \draw (521,90) node [anchor=north west][inner sep=0.75pt]  [font=\small] [align=left] {DS};
        % Text Node
        \draw (230,96.67) node [anchor=north west][inner sep=0.75pt]  [font=\footnotesize] [align=left] {1};
        % Text Node
        \draw (230,113.43) node [anchor=north west][inner sep=0.75pt]  [font=\footnotesize] [align=left] {2};
        % Text Node
        \draw (230,130.19) node [anchor=north west][inner sep=0.75pt]  [font=\footnotesize] [align=left] {3};
        % Text Node
        \draw (230,146.95) node [anchor=north west][inner sep=0.75pt]  [font=\footnotesize] [align=left] {4};
        % Text Node
        \draw (230,163.71) node [anchor=north west][inner sep=0.75pt]  [font=\footnotesize] [align=left] {5};
        % Text Node
        \draw (230,180.47) node [anchor=north west][inner sep=0.75pt]  [font=\footnotesize] [align=left] {6};
        % Text Node
        \draw (230,197.23) node [anchor=north west][inner sep=0.75pt]  [font=\footnotesize] [align=left] {7};
        % Text Node
        \draw (230,214) node [anchor=north west][inner sep=0.75pt]  [font=\footnotesize] [align=left] {8};
        % Text Node
        \draw (242,82.17) node [anchor=north west][inner sep=0.75pt]  [font=\footnotesize] [align=left] {1};
        % Text Node
        \draw (256,82.17) node [anchor=north west][inner sep=0.75pt]  [font=\footnotesize] [align=left] {2};
        % Text Node
        \draw (272,82.17) node [anchor=north west][inner sep=0.75pt]  [font=\footnotesize] [align=left] {3};
        % Text Node
        \draw (287,82.17) node [anchor=north west][inner sep=0.75pt]  [font=\footnotesize] [align=left] {4};
        % Text Node
        \draw (571.6,251.4) node [anchor=north west][inner sep=0.75pt]    {$\hat{\mathbf{x}}_{1:T}^{4} ,p_{1:T}^{4} ,0$};

        \end{tikzpicture}
    }
    \caption{Overview of the DDA+DS method. A scene containing a sequence of measurements is processed by the DDA, which produces a data association matrix. This is then used for partitioning (represented by the box P) the measurement sequence, and each partition is individually fed to the deep smoother module for predicting a track.}
    \label{fig:overview}

\end{figure*}
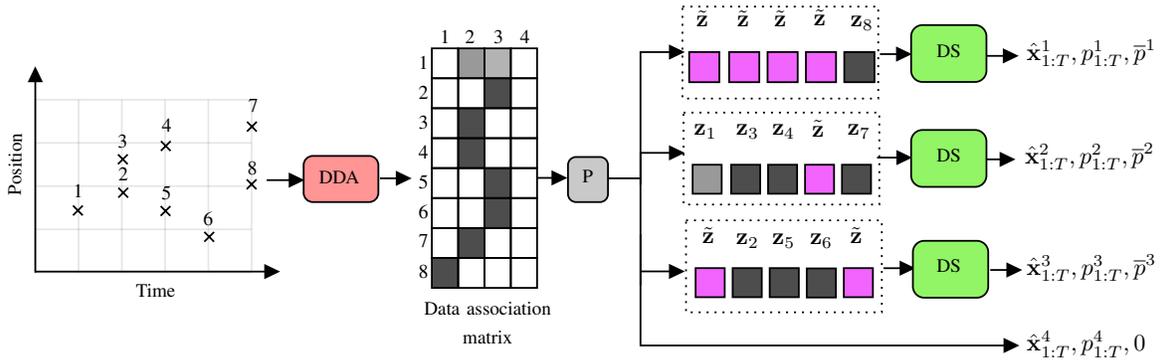

Figure \ref{fig:overview} shows an example of the method processing a measurement sequence into trajectories. On the left of the image, a measurement sequence $\mathbf z_{1:n}$ with eight measurements is shown, which is fed to the DDA module. This module attends to all measurements simultaneously and predicts the data association matrix $\mathbf A\in\mathbb R^{n\times B}$, where each row corresponds to one probability mass function (pmf) over $B$ possible tracks. Each column of the matrix corresponds to one track, where $B=4$ in this example. This matrix can be interpreted as a soft, trainable association between measurements and tracks.
Connecting the DDA's module output to the next module's input is a partitioning step, which uses the predicted data association matrix to partition the measurements into $B$ different tracks. This step also discards tracks that do not have any associated measurements (e.g., track 4 in Fig.\,\ref{fig:overview}).%by setting their $\bar{p}=0$ and arbitrary values to $\hat{\mathbf x}_{1:T}$ and $p_{1:T}$. 
Each remaining track is then individually fed to the DS module, shown in green on the right, which processes the measurements in each track into the corresponding parameters $(\hat{\mathbf x}_{1:T}, p_{1:T}, \bar{p})$ of its predicted trajectory.

Decoupling the data association from the smoothing task brings many benefits. Evidently, it increases the interpretability of the model by providing access to the data associations used, allowing users to better understand the predictions and diagnose, e.g., if a bad prediction was due to incorrect data associations or because the smoothing process was not able to provide reasonable estimates. However, added interpretability is not the only advantage. Many transformer-based multi-object trackers rely on the transformer decoder to produce estimates, but the involved cross-attention and object query training have been associated with slow convergence \cite{rethinking_transformer_set_prediction, dn_detr, dino_detr, dab_detr, deformable-DETR}. By decoupling the data association from smoothing, D3AS sidesteps the need for a decoder, which reduces training time, inference time, and model size.

\subsection{Deep Data Association Module}
%In order to perform data association in a differentiable manner, we propose a structure that predicts soft associations between measurements and tracks via a data association matrix $\mathbf A$, as shown in Fig.\,\ref{fig:overview}. Note how the data association matrix predicts three likely tracks in the illustrated example: one track containing measurements 1, 3, 4, and 7, another track containing measurements 2, 5 and 6, and one last track with only measurement 8 in it. At the same time, it also displays that the association for measurement 1 is uncertain: although it is more likely to belong to track 2, it is also possible for it to belong to track 3.

\begin{figure}[t]
    \centering

\tikzset{every picture/.style={line width=0.75pt}} %set default line width to 0.75pt        

\begin{tikzpicture}[x=0.75pt,y=0.75pt,yscale=-1,xscale=1]
%uncomment if require: \path (0,322); %set diagram left start at 0, and has height of 322

%Rounded Rect [id:dp05234236909233658] 
\draw  [fill={rgb, 255:red, 255; green, 177; blue, 178 }  ,fill opacity=1 ][dash pattern={on 4.5pt off 4.5pt}] (11.9,101.26) .. controls (11.9,83.66) and (26.17,69.39) .. (43.77,69.39) -- (180.53,69.39) .. controls (198.13,69.39) and (212.4,83.66) .. (212.4,101.26) -- (212.4,196.86) .. controls (212.4,214.46) and (198.13,228.72) .. (180.53,228.72) -- (43.77,228.72) .. controls (26.17,228.72) and (11.9,214.46) .. (11.9,196.86) -- cycle ;
%Straight Lines [id:da4523267524893806] 
\draw    (47,234.06) -- (47,214.74) ;
\draw [shift={(47,211.74)}, rotate = 90] [fill={rgb, 255:red, 0; green, 0; blue, 0 }  ][line width=0.08]  [draw opacity=0] (8.93,-4.29) -- (0,0) -- (8.93,4.29) -- cycle    ;
%Straight Lines [id:da7008973438236885] 
\draw    (94,234.06) -- (94,214.74) ;
\draw [shift={(94,211.74)}, rotate = 90] [fill={rgb, 255:red, 0; green, 0; blue, 0 }  ][line width=0.08]  [draw opacity=0] (8.93,-4.29) -- (0,0) -- (8.93,4.29) -- cycle    ;
%Straight Lines [id:da29777661102069275] 
\draw    (177,233.31) -- (177,214) ;
\draw [shift={(177,211)}, rotate = 90] [fill={rgb, 255:red, 0; green, 0; blue, 0 }  ][line width=0.08]  [draw opacity=0] (8.93,-4.29) -- (0,0) -- (8.93,4.29) -- cycle    ;
%Straight Lines [id:da4533248959134417] 
\draw    (47,178) -- (47,159.12) ;
\draw [shift={(47,156.12)}, rotate = 90] [fill={rgb, 255:red, 0; green, 0; blue, 0 }  ][line width=0.08]  [draw opacity=0] (8.93,-4.29) -- (0,0) -- (8.93,4.29) -- cycle    ;
%Straight Lines [id:da0746150222771007] 
\draw    (94,178) -- (94,159.12) ;
\draw [shift={(94,156.12)}, rotate = 90] [fill={rgb, 255:red, 0; green, 0; blue, 0 }  ][line width=0.08]  [draw opacity=0] (8.93,-4.29) -- (0,0) -- (8.93,4.29) -- cycle    ;
%Straight Lines [id:da3832107805915854] 
\draw    (177,177.27) -- (177,158.39) ;
\draw [shift={(177,155.39)}, rotate = 90] [fill={rgb, 255:red, 0; green, 0; blue, 0 }  ][line width=0.08]  [draw opacity=0] (8.93,-4.29) -- (0,0) -- (8.93,4.29) -- cycle    ;
%Rounded Rect [id:dp9296968385466307] 
\draw  [color={rgb, 255:red, 0; green, 0; blue, 0 }  ,draw opacity=1 ][fill={rgb, 255:red, 252; green, 157; blue, 255 }  ,fill opacity=1 ] (158,82.85) .. controls (158,79.76) and (160.51,77.25) .. (163.6,77.25) -- (188.4,77.25) .. controls (191.49,77.25) and (194,79.76) .. (194,82.85) -- (194,99.65) .. controls (194,102.74) and (191.49,105.25) .. (188.4,105.25) -- (163.6,105.25) .. controls (160.51,105.25) and (158,102.74) .. (158,99.65) -- cycle ;
%Rounded Rect [id:dp7159639762648593] 
\draw  [color={rgb, 255:red, 0; green, 0; blue, 0 }  ,draw opacity=1 ][fill={rgb, 255:red, 252; green, 157; blue, 255 }  ,fill opacity=1 ] (75.5,82.85) .. controls (75.5,79.76) and (78.01,77.25) .. (81.1,77.25) -- (105.9,77.25) .. controls (108.99,77.25) and (111.5,79.76) .. (111.5,82.85) -- (111.5,99.65) .. controls (111.5,102.74) and (108.99,105.25) .. (105.9,105.25) -- (81.1,105.25) .. controls (78.01,105.25) and (75.5,102.74) .. (75.5,99.65) -- cycle ;
%Rounded Rect [id:dp9819240432356504] 
\draw  [color={rgb, 255:red, 0; green, 0; blue, 0 }  ,draw opacity=1 ][fill={rgb, 255:red, 252; green, 157; blue, 255 }  ,fill opacity=1 ] (28.5,82.85) .. controls (28.5,79.76) and (31.01,77.25) .. (34.1,77.25) -- (58.9,77.25) .. controls (61.99,77.25) and (64.5,79.76) .. (64.5,82.85) -- (64.5,99.65) .. controls (64.5,102.74) and (61.99,105.25) .. (58.9,105.25) -- (34.1,105.25) .. controls (31.01,105.25) and (28.5,102.74) .. (28.5,99.65) -- cycle ;
%Rounded Rect [id:dp6281538395813133] 
\draw  [color={rgb, 255:red, 0; green, 0; blue, 0 }  ,draw opacity=1 ][fill={rgb, 255:red, 12; green, 252; blue, 234 }  ,fill opacity=1 ] (28.5,133.35) .. controls (28.5,130.26) and (31.01,127.75) .. (34.1,127.75) -- (58.9,127.75) .. controls (61.99,127.75) and (64.5,130.26) .. (64.5,133.35) -- (64.5,150.15) .. controls (64.5,153.24) and (61.99,155.75) .. (58.9,155.75) -- (34.1,155.75) .. controls (31.01,155.75) and (28.5,153.24) .. (28.5,150.15) -- cycle ;
%Rounded Rect [id:dp22060031256878654] 
\draw  [color={rgb, 255:red, 0; green, 0; blue, 0 }  ,draw opacity=1 ][fill={rgb, 255:red, 12; green, 252; blue, 234 }  ,fill opacity=1 ] (76,133.35) .. controls (76,130.26) and (78.51,127.75) .. (81.6,127.75) -- (106.4,127.75) .. controls (109.49,127.75) and (112,130.26) .. (112,133.35) -- (112,150.15) .. controls (112,153.24) and (109.49,155.75) .. (106.4,155.75) -- (81.6,155.75) .. controls (78.51,155.75) and (76,153.24) .. (76,150.15) -- cycle ;
%Rounded Rect [id:dp7358871799298237] 
\draw  [color={rgb, 255:red, 0; green, 0; blue, 0 }  ,draw opacity=1 ][fill={rgb, 255:red, 12; green, 252; blue, 234 }  ,fill opacity=1 ] (158.5,132.35) .. controls (158.5,129.26) and (161.01,126.75) .. (164.1,126.75) -- (188.9,126.75) .. controls (191.99,126.75) and (194.5,129.26) .. (194.5,132.35) -- (194.5,149.15) .. controls (194.5,152.24) and (191.99,154.75) .. (188.9,154.75) -- (164.1,154.75) .. controls (161.01,154.75) and (158.5,152.24) .. (158.5,149.15) -- cycle ;
%Rounded Rect [id:dp729767081842241] 
\draw  [fill={rgb, 255:red, 255; green, 201; blue, 32 }  ,fill opacity=1 ] (34.71,184.15) .. controls (34.71,180.44) and (37.72,177.44) .. (41.43,177.44) -- (184,177.44) .. controls (187.71,177.44) and (190.71,180.44) .. (190.71,184.15) -- (190.71,204.29) .. controls (190.71,207.99) and (187.71,211) .. (184,211) -- (41.43,211) .. controls (37.72,211) and (34.71,207.99) .. (34.71,204.29) -- cycle ;
%Straight Lines [id:da29788192712647965] 
\draw    (219,194.6) -- (194.4,194.6) ;
\draw [shift={(191.4,194.6)}, rotate = 360] [fill={rgb, 255:red, 0; green, 0; blue, 0 }  ][line width=0.08]  [draw opacity=0] (8.93,-4.29) -- (0,0) -- (8.93,4.29) -- cycle    ;
%Straight Lines [id:da2873011673668925] 
\draw    (46.33,127.67) -- (46.33,108.78) ;
\draw [shift={(46.33,105.78)}, rotate = 90] [fill={rgb, 255:red, 0; green, 0; blue, 0 }  ][line width=0.08]  [draw opacity=0] (8.93,-4.29) -- (0,0) -- (8.93,4.29) -- cycle    ;
%Straight Lines [id:da28675046489075806] 
\draw    (93.33,127.67) -- (93.33,108.78) ;
\draw [shift={(93.33,105.78)}, rotate = 90] [fill={rgb, 255:red, 0; green, 0; blue, 0 }  ][line width=0.08]  [draw opacity=0] (8.93,-4.29) -- (0,0) -- (8.93,4.29) -- cycle    ;
%Straight Lines [id:da8852363611151336] 
\draw    (176.33,126.94) -- (176.33,108.06) ;
\draw [shift={(176.33,105.06)}, rotate = 90] [fill={rgb, 255:red, 0; green, 0; blue, 0 }  ][line width=0.08]  [draw opacity=0] (8.93,-4.29) -- (0,0) -- (8.93,4.29) -- cycle    ;
%Straight Lines [id:da024117769825365842] 
\draw    (47,77.67) -- (47,58.78) ;
\draw [shift={(47,55.78)}, rotate = 90] [fill={rgb, 255:red, 0; green, 0; blue, 0 }  ][line width=0.08]  [draw opacity=0] (8.93,-4.29) -- (0,0) -- (8.93,4.29) -- cycle    ;
%Straight Lines [id:da14656049063502874] 
\draw    (94,77.67) -- (94,58.78) ;
\draw [shift={(94,55.78)}, rotate = 90] [fill={rgb, 255:red, 0; green, 0; blue, 0 }  ][line width=0.08]  [draw opacity=0] (8.93,-4.29) -- (0,0) -- (8.93,4.29) -- cycle    ;
%Straight Lines [id:da8982787595522351] 
\draw    (177,76.94) -- (177,58.06) ;
\draw [shift={(177,55.06)}, rotate = 90] [fill={rgb, 255:red, 0; green, 0; blue, 0 }  ][line width=0.08]  [draw opacity=0] (8.93,-4.29) -- (0,0) -- (8.93,4.29) -- cycle    ;

% Text Node
\draw (40,235.9) node [anchor=north west][inner sep=0.75pt]    {$\mathbf{z}_{1}$};
% Text Node
\draw (93,188) node [anchor=north west][inner sep=0.75pt]  [font=\small] [align=left] {Encoder};
% Text Node
\draw (34,135) node [anchor=north west][inner sep=0.75pt]  [font=\small] [align=left] {FFN};
% Text Node
\draw (170,235.9) node [anchor=north west][inner sep=0.75pt]    {$\mathbf{z}_{n}$};
% Text Node
\draw (125,236.9) node [anchor=north west][inner sep=0.75pt]    {$\cdots $};
% Text Node
\draw (87,235.9) node [anchor=north west][inner sep=0.75pt]    {$\mathbf{z}_{2}$};
% Text Node
\draw (81,135) node [anchor=north west][inner sep=0.75pt]  [font=\small] [align=left] {FFN};
% Text Node
\draw (164,134) node [anchor=north west][inner sep=0.75pt]  [font=\small] [align=left] {FFN};
% Text Node
\draw (124,136) node [anchor=north west][inner sep=0.75pt]    {$\cdots $};
% Text Node
\draw (36,85) node [anchor=north west][inner sep=0.75pt]  [font=\small] [align=left] {SM};
% Text Node
\draw (124,87) node [anchor=north west][inner sep=0.75pt]    {$\cdots $};
% Text Node
\draw (84,85) node [anchor=north west][inner sep=0.75pt]  [font=\small] [align=left] {SM};
% Text Node
\draw (166,85) node [anchor=north west][inner sep=0.75pt]  [font=\small] [align=left] {SM};
% Text Node
\draw (38,42.07) node [anchor=north west][inner sep=0.75pt]    {$\mathbf{y}_{1}$};
% Text Node
\draw (85,42.07) node [anchor=north west][inner sep=0.75pt]    {$\mathbf{y}_{2}$};
% Text Node
\draw (167,42.07) node [anchor=north west][inner sep=0.75pt]    {$\mathbf{y}_{n}$};
% Text Node
\draw (124,46.07) node [anchor=north west][inner sep=0.75pt]    {$\cdots $};
% Text Node
\draw (221,190) node [anchor=north west][inner sep=0.75pt]    {$\mathbf{q}_{1:n}$};

\end{tikzpicture}

    \caption{Structure of the Deep Data Associator module. A sequence of measurements $\mathbf z_{1:n}$ is processed by a transformer encoder that uses temporal encodings, followed by an individual application of FFN and Softmax layers to the computed embedding for each measurement.}
    \label{fig:dda}
    \vspace{-10pt}
\end{figure}
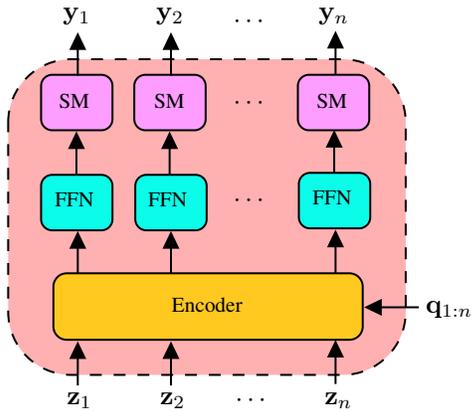

The high-level functioning of the DDA module is shown in Fig.\,\ref{fig:dda}. First, the measurements in the set $\mathbb Z_\Omega$ are put in arbitrary order in a sequence $\mathbf z_{1:n}$, and the corresponding times-of-arrival in the same order are put into another sequence $t_{1:n}$. Then, we project each element of the sequence $\mathbf z_i$ into a higher-dimensional representation $\mathbf z'_i\in\mathbb R^{d'}$ with the help of an FFN layer\footnote{The projection of z into z' is not shown in Fig. 3 to avoid clutter.}, and use a transformer encoder to process this new measurement sequence into embeddings $\mathbf e_{1:n}$. 

For the encoder to know \emph{when} each measurement was obtained, we use positional encodings based on the time-of-arrival of the measurements, as introduced in \eqref{eq:TOA}. In specific, the positional encoding $\mathbf q_i$ used by the DDA's encoder for element $i$ of the input sequence is based on the time-of-arrival for that measurement: $\mathbf q_i = f(t_i)$,
%\begin{equation}
%    \mathbf q_{1:n} = \left[~f(t_1), ~\cdots,~ f(t_n)~\right],
%\end{equation}
where $f$ is a learnable lookup table trained jointly with the other DDA's learnable parameters. 

Using a transformer encoder allows the DDA to efficiently learn complex temporal dependencies between all the measurements, enabling powerful data association predictions conditioned on the entire set $\mathbb Z_\Omega$. After the encoder, each embedding is individually fed to a feedforward neural network (FFN) and a Softmax (SM) layer, resulting in a pmf $\mathbf y_i$ for each measurement $\mathbf z_i$, which are used to form the rows of $\mathbf A$, i.e., $\mathbf A =
\begin{bmatrix}
    \mathbf y_1^\top & \cdots &  \mathbf y_n^\top
\end{bmatrix}^\top$. 
The DDA module is fully trainable, and we propose a novel loss to allow supervised training of the prediction $\mathbf A$, described in subsection \ref{subsec:losses}.

\subsection{Partitioning}
The partitioning step in D3AS uses the predicted data association matrix to decide how to partition the measurement sequence $\mathbf z_{1:n}$ before feeding it to the DS module. We find that a simple greedy assignment performs reasonably well in the cases studied, compared to the model-based benchmark: each measurement is associated to the track corresponding to the mode of its predicted pmf, and tracks that end up without any measurements are discarded. To provide a proxy for how much the smoother should trust each measurement, we concatenate the maximum value of each measurement's association pmf to the measurement vector. 
Further, by using the time-of-arrival information of each measurement, we sort them in arrival order into a sequence and add a dummy vector $\tilde{\mathbf z}$ to all the time-steps where a track does not have measurements. If multiple measurements with the same time-of-arrival end up in the same track, we remove all but one of them. These two steps enforce all tracks to have exactly $T$ elements, regardless of how many measurements that track contains.

A sample output of this process is illustrated in Fig.\,\ref{fig:overview}, showing the partitioning of the measurements into four different tracks. Time-steps with missing measurements (e.g., the first four time-steps for track 1) use a dummy $\tilde{\mathbf z}$ vector, and the DS module will be given the information that measurement $\mathbf z_1$ is less certain than the others (illustrated by making the box underneath it less opaque).

\subsection{Deep Smoother Module}
\begin{figure}[t]
    \centering

\tikzset{every picture/.style={line width=0.75pt}} %set default line width to 0.75pt        

\begin{tikzpicture}[x=0.75pt,y=0.75pt,yscale=-1,xscale=1]
%uncomment if require: \path (0,371); %set diagram left start at 0, and has height of 371

%Rounded Rect [id:dp9819592886577615] 
\draw  [fill={rgb, 255:red, 193; green, 254; blue, 179 }  ,fill opacity=1 ][dash pattern={on 4.5pt off 4.5pt}] (107.8,170.06) .. controls (107.8,157.54) and (117.95,147.39) .. (130.47,147.39) -- (317.13,147.39) .. controls (329.65,147.39) and (339.8,157.54) .. (339.8,170.06) -- (339.8,238.06) .. controls (339.8,250.57) and (329.65,260.72) .. (317.13,260.72) -- (130.47,260.72) .. controls (117.95,260.72) and (107.8,250.57) .. (107.8,238.06) -- cycle ;
%Shape: Rectangle [id:dp5663151470992396] 
\draw  [fill={rgb, 255:red, 0; green, 0; blue, 0 }  ,fill opacity=0.4 ] (124.96,292.01) -- (143.82,292.01) -- (143.82,310.87) -- (124.96,310.87) -- cycle ;
%Shape: Rectangle [id:dp3686934821192953] 
\draw  [fill={rgb, 255:red, 0; green, 0; blue, 0 }  ,fill opacity=0.7 ] (161.31,292.01) -- (180.17,292.01) -- (180.17,310.87) -- (161.31,310.87) -- cycle ;
%Shape: Rectangle [id:dp6016037370108904] 
\draw  [fill={rgb, 255:red, 0; green, 0; blue, 0 }  ,fill opacity=0.7 ] (196.06,292.01) -- (214.91,292.01) -- (214.91,310.87) -- (196.06,310.87) -- cycle ;
%Shape: Rectangle [id:dp1943003130561931] 
\draw  [fill={rgb, 255:red, 0; green, 0; blue, 0 }  ,fill opacity=0.7 ] (262.96,292.01) -- (281.82,292.01) -- (281.82,310.87) -- (262.96,310.87) -- cycle ;
%Shape: Rectangle [id:dp6334527775593006] 
\draw  [fill={rgb, 255:red, 235; green, 34; blue, 254 }  ,fill opacity=0.7 ] (229.4,292.01) -- (248.26,292.01) -- (248.26,310.87) -- (229.4,310.87) -- cycle ;
%Straight Lines [id:da8268388892798706] 
\draw    (313.8,174.2) -- (313.8,132.75) ;
\draw [shift={(313.8,129.75)}, rotate = 90] [fill={rgb, 255:red, 0; green, 0; blue, 0 }  ][line width=0.08]  [draw opacity=0] (8.93,-4.29) -- (0,0) -- (8.93,4.29) -- cycle    ;
%Shape: Rectangle [id:dp5694010383615127] 
\draw  [dash pattern={on 0.84pt off 2.51pt}] (120.11,270.72) -- (288.78,270.72) -- (288.78,316.06) -- (120.11,316.06) -- cycle ;
%Rounded Rect [id:dp5643404505182426] 
\draw  [fill={rgb, 255:red, 255; green, 201; blue, 32 }  ,fill opacity=1 ] (117,213.35) .. controls (117,209.64) and (120.01,206.64) .. (123.71,206.64) -- (289.49,206.64) .. controls (293.19,206.64) and (296.2,209.64) .. (296.2,213.35) -- (296.2,233.49) .. controls (296.2,237.19) and (293.19,240.2) .. (289.49,240.2) -- (123.71,240.2) .. controls (120.01,240.2) and (117,237.19) .. (117,233.49) -- cycle ;
%Rounded Rect [id:dp11937643455371338] 
\draw  [color={rgb, 255:red, 0; green, 0; blue, 0 }  ,draw opacity=1 ][fill={rgb, 255:red, 12; green, 252; blue, 234 }  ,fill opacity=1 ] (295.9,180.55) .. controls (295.9,177.46) and (298.41,174.95) .. (301.5,174.95) -- (326.3,174.95) .. controls (329.39,174.95) and (331.9,177.46) .. (331.9,180.55) -- (331.9,197.35) .. controls (331.9,200.44) and (329.39,202.95) .. (326.3,202.95) -- (301.5,202.95) .. controls (298.41,202.95) and (295.9,200.44) .. (295.9,197.35) -- cycle ;

%Straight Lines [id:da027326222357480834] 
\draw    (41.2,223.8) -- (67,223.8) -- (67,208.27) ;
\draw [shift={(67,205.27)}, rotate = 90] [fill={rgb, 255:red, 0; green, 0; blue, 0 }  ][line width=0.08]  [draw opacity=0] (8.93,-4.29) -- (0,0) -- (8.93,4.29) -- cycle    ;
%Straight Lines [id:da5456098359032937] 
\draw    (28,207.47) -- (28,223.67) -- (50.2,223.8) -- (50.2,240.2) ;
\draw [shift={(28,204.47)}, rotate = 90] [fill={rgb, 255:red, 0; green, 0; blue, 0 }  ][line width=0.08]  [draw opacity=0] (8.93,-4.29) -- (0,0) -- (8.93,4.29) -- cycle    ;
%Straight Lines [id:da4918361831140641] 
\draw    (67.6,177.8) -- (67.6,167.8) ;
%Straight Lines [id:da9510626817305816] 
\draw    (27.6,175.8) -- (27.6,118.8) ;
\draw [shift={(27.6,115.8)}, rotate = 90] [fill={rgb, 255:red, 0; green, 0; blue, 0 }  ][line width=0.08]  [draw opacity=0] (8.93,-4.29) -- (0,0) -- (8.93,4.29) -- cycle    ;
%Rounded Rect [id:dp7018244065899215] 
\draw  [color={rgb, 255:red, 0; green, 0; blue, 0 }  ,draw opacity=1 ][fill={rgb, 255:red, 12; green, 252; blue, 234 }  ,fill opacity=1 ] (9.7,182.65) .. controls (9.7,179.56) and (12.21,177.05) .. (15.3,177.05) -- (40.1,177.05) .. controls (43.19,177.05) and (45.7,179.56) .. (45.7,182.65) -- (45.7,199.45) .. controls (45.7,202.54) and (43.19,205.05) .. (40.1,205.05) -- (15.3,205.05) .. controls (12.21,205.05) and (9.7,202.54) .. (9.7,199.45) -- cycle ;

%Rounded Rect [id:dp2533324356387452] 
\draw  [color={rgb, 255:red, 0; green, 0; blue, 0 }  ,draw opacity=1 ][fill={rgb, 255:red, 12; green, 252; blue, 234 }  ,fill opacity=1 ] (49.7,182.65) .. controls (49.7,179.56) and (52.21,177.05) .. (55.3,177.05) -- (80.1,177.05) .. controls (83.19,177.05) and (85.7,179.56) .. (85.7,182.65) -- (85.7,199.45) .. controls (85.7,202.54) and (83.19,205.05) .. (80.1,205.05) -- (55.3,205.05) .. controls (52.21,205.05) and (49.7,202.54) .. (49.7,199.45) -- cycle ;

%Rounded Rect [id:dp2542605086521599] 
\draw  [color={rgb, 255:red, 0; green, 0; blue, 0 }  ,draw opacity=1 ][fill={rgb, 255:red, 252; green, 157; blue, 255 }  ,fill opacity=1 ] (49.2,144.65) .. controls (49.2,141.56) and (51.71,139.05) .. (54.8,139.05) -- (79.6,139.05) .. controls (82.69,139.05) and (85.2,141.56) .. (85.2,144.65) -- (85.2,161.45) .. controls (85.2,164.54) and (82.69,167.05) .. (79.6,167.05) -- (54.8,167.05) .. controls (51.71,167.05) and (49.2,164.54) .. (49.2,161.45) -- cycle ;

%Straight Lines [id:da2751154023642304] 
\draw    (67.4,138.53) -- (67.4,119.2) ;
\draw [shift={(67.4,116.2)}, rotate = 90] [fill={rgb, 255:red, 0; green, 0; blue, 0 }  ][line width=0.08]  [draw opacity=0] (8.93,-4.29) -- (0,0) -- (8.93,4.29) -- cycle    ;
%Rounded Rect [id:dp5116499631147817] 
\draw  [dash pattern={on 4.5pt off 4.5pt}] (4.8,147.05) .. controls (4.8,137.33) and (12.68,129.45) .. (22.4,129.45) -- (75.2,129.45) .. controls (84.92,129.45) and (92.8,137.33) .. (92.8,147.05) -- (92.8,214.6) .. controls (92.8,224.32) and (84.92,232.2) .. (75.2,232.2) -- (22.4,232.2) .. controls (12.68,232.2) and (4.8,224.32) .. (4.8,214.6) -- cycle ;
%Straight Lines [id:da09742578399102508] 
\draw    (135,206.64) -- (135,185.47) ;
\draw [shift={(135,182.47)}, rotate = 90] [fill={rgb, 255:red, 0; green, 0; blue, 0 }  ][line width=0.08]  [draw opacity=0] (8.93,-4.29) -- (0,0) -- (8.93,4.29) -- cycle    ;
%Straight Lines [id:da11702109049280418] 
\draw    (169.62,206.64) -- (169.62,185.47) ;
\draw [shift={(169.62,182.47)}, rotate = 90] [fill={rgb, 255:red, 0; green, 0; blue, 0 }  ][line width=0.08]  [draw opacity=0] (8.93,-4.29) -- (0,0) -- (8.93,4.29) -- cycle    ;
%Straight Lines [id:da24119198573372636] 
\draw    (272.49,206.72) -- (272.49,185.55) ;
\draw [shift={(272.49,182.55)}, rotate = 90] [fill={rgb, 255:red, 0; green, 0; blue, 0 }  ][line width=0.08]  [draw opacity=0] (8.93,-4.29) -- (0,0) -- (8.93,4.29) -- cycle    ;
%Straight Lines [id:da5399258089441983] 
\draw    (134.4,270.13) -- (134.4,244.07) ;
\draw [shift={(134.4,241.07)}, rotate = 90] [fill={rgb, 255:red, 0; green, 0; blue, 0 }  ][line width=0.08]  [draw opacity=0] (8.93,-4.29) -- (0,0) -- (8.93,4.29) -- cycle    ;
%Straight Lines [id:da08184303162569329] 
\draw    (169.02,270.13) -- (169.02,244.07) ;
\draw [shift={(169.02,241.07)}, rotate = 90] [fill={rgb, 255:red, 0; green, 0; blue, 0 }  ][line width=0.08]  [draw opacity=0] (8.93,-4.29) -- (0,0) -- (8.93,4.29) -- cycle    ;
%Straight Lines [id:da4398403085400091] 
\draw    (203.64,270.13) -- (203.64,244.07) ;
\draw [shift={(203.64,241.07)}, rotate = 90] [fill={rgb, 255:red, 0; green, 0; blue, 0 }  ][line width=0.08]  [draw opacity=0] (8.93,-4.29) -- (0,0) -- (8.93,4.29) -- cycle    ;
%Straight Lines [id:da5629200881544429] 
\draw    (238.26,270.13) -- (238.26,244.07) ;
\draw [shift={(238.26,241.07)}, rotate = 90] [fill={rgb, 255:red, 0; green, 0; blue, 0 }  ][line width=0.08]  [draw opacity=0] (8.93,-4.29) -- (0,0) -- (8.93,4.29) -- cycle    ;
%Straight Lines [id:da19242701761014258] 
\draw    (272.89,270.24) -- (272.89,244.17) ;
\draw [shift={(272.89,241.17)}, rotate = 90] [fill={rgb, 255:red, 0; green, 0; blue, 0 }  ][line width=0.08]  [draw opacity=0] (8.93,-4.29) -- (0,0) -- (8.93,4.29) -- cycle    ;
%Straight Lines [id:da659702264040863] 
\draw  [dash pattern={on 4.5pt off 4.5pt}]  (86.11,132.72) -- (134.4,154.6) ;
%Straight Lines [id:da7529191410629645] 
\draw  [dash pattern={on 4.5pt off 4.5pt}]  (90.78,222.72) -- (135,182.47) ;
%Straight Lines [id:da7055276105461679] 
\draw    (134.4,159.08) -- (134.4,131.72) ;
\draw [shift={(134.4,128.72)}, rotate = 90] [fill={rgb, 255:red, 0; green, 0; blue, 0 }  ][line width=0.08]  [draw opacity=0] (8.93,-4.29) -- (0,0) -- (8.93,4.29) -- cycle    ;
%Straight Lines [id:da7613040625194474] 
\draw    (169.6,159.08) -- (169.6,131.72) ;
\draw [shift={(169.6,128.72)}, rotate = 90] [fill={rgb, 255:red, 0; green, 0; blue, 0 }  ][line width=0.08]  [draw opacity=0] (8.93,-4.29) -- (0,0) -- (8.93,4.29) -- cycle    ;
%Straight Lines [id:da124880969867462] 
\draw    (272,159.85) -- (272,132.49) ;
\draw [shift={(272,129.49)}, rotate = 90] [fill={rgb, 255:red, 0; green, 0; blue, 0 }  ][line width=0.08]  [draw opacity=0] (8.93,-4.29) -- (0,0) -- (8.93,4.29) -- cycle    ;
%Straight Lines [id:da0728110436631586] 
\draw    (289.44,286.47) -- (312.6,286.47) -- (312.6,206.39) ;
\draw [shift={(312.6,203.39)}, rotate = 90] [fill={rgb, 255:red, 0; green, 0; blue, 0 }  ][line width=0.08]  [draw opacity=0] (8.93,-4.29) -- (0,0) -- (8.93,4.29) -- cycle    ;
%Rounded Rect [id:dp531623468610178] 
\draw  [color={rgb, 255:red, 0; green, 0; blue, 0 }  ,draw opacity=1 ][fill={rgb, 255:red, 182; green, 177; blue, 255 }  ,fill opacity=1 ] (157.1,159.65) .. controls (157.1,156.94) and (159.29,154.75) .. (162,154.75) -- (176.7,154.75) .. controls (179.41,154.75) and (181.6,156.94) .. (181.6,159.65) -- (181.6,177.85) .. controls (181.6,180.56) and (179.41,182.75) .. (176.7,182.75) -- (162,182.75) .. controls (159.29,182.75) and (157.1,180.56) .. (157.1,177.85) -- cycle ;
%Rounded Rect [id:dp9005962272733338] 
\draw  [color={rgb, 255:red, 0; green, 0; blue, 0 }  ,draw opacity=1 ][fill={rgb, 255:red, 182; green, 177; blue, 255 }  ,fill opacity=1 ] (122.7,159.45) .. controls (122.7,156.74) and (124.89,154.55) .. (127.6,154.55) -- (142.3,154.55) .. controls (145.01,154.55) and (147.2,156.74) .. (147.2,159.45) -- (147.2,177.65) .. controls (147.2,180.36) and (145.01,182.55) .. (142.3,182.55) -- (127.6,182.55) .. controls (124.89,182.55) and (122.7,180.36) .. (122.7,177.65) -- cycle ;
%Rounded Rect [id:dp4824370002135714] 
\draw  [color={rgb, 255:red, 0; green, 0; blue, 0 }  ,draw opacity=1 ][fill={rgb, 255:red, 182; green, 177; blue, 255 }  ,fill opacity=1 ] (260.5,159.65) .. controls (260.5,156.94) and (262.69,154.75) .. (265.4,154.75) -- (280.1,154.75) .. controls (282.81,154.75) and (285,156.94) .. (285,159.65) -- (285,177.85) .. controls (285,180.56) and (282.81,182.75) .. (280.1,182.75) -- (265.4,182.75) .. controls (262.69,182.75) and (260.5,180.56) .. (260.5,177.85) -- cycle ;

% Text Node
\draw (126.51,277.2) node [anchor=north west][inner sep=0.75pt]    {$\mathbf{z}_{1}$};
% Text Node
\draw (163.47,277.2) node [anchor=north west][inner sep=0.75pt]    {$\mathbf{z}_{3}$};
% Text Node
\draw (197.83,277.2) node [anchor=north west][inner sep=0.75pt]    {$\mathbf{z}_{4}$};
% Text Node
\draw (265.2,277.2) node [anchor=north west][inner sep=0.75pt]    {$\mathbf{z}_{7}$};
% Text Node
\draw (233.43,274.8) node [anchor=north west][inner sep=0.75pt]    {$\tilde{\mathbf{z}}_\text{DS}$};
% Text Node
\draw (183,216) node [anchor=north west][inner sep=0.75pt]  [color={rgb, 255:red, 0; green, 0; blue, 0 }  ,opacity=1 ] [align=left] {Encoder};
% Text Node
\draw (309.2,116.1) node [anchor=north west][inner sep=0.75pt]  [font=\small]  {$\overline{p}$};
% Text Node
\draw (300.9,182) node [anchor=north west][inner sep=0.75pt]  [font=\small] [align=left] {FFN};
% Text Node
\draw (115.4,115.19) node [anchor=north west][inner sep=0.75pt]  [font=\footnotesize]  {$\hat{\mathbf{x}}_{1} ,p_{1}$};
% Text Node
\draw (196,156.2) node [anchor=north west][inner sep=0.75pt]  [font=\Large]  {$\cdots $};
% Text Node
\draw (56.7,147) node [anchor=north west][inner sep=0.75pt]  [font=\small] [align=left] {SM};
% Text Node
\draw (55,185) node [anchor=north west][inner sep=0.75pt]  [font=\small] [align=left] {FFN};
% Text Node
\draw (15,185) node [anchor=north west][inner sep=0.75pt]  [font=\small] [align=left] {FFN};
% Text Node
\draw (156.4,115.56) node [anchor=north west][inner sep=0.75pt]  [font=\footnotesize]  {$\hat{\mathbf{x}}_{2} ,p_{2}$};
% Text Node
\draw (255.8,115.56) node [anchor=north west][inner sep=0.75pt]  [font=\footnotesize]  {$\hat{\mathbf{x}}_{5} ,p_{5}$};
% Text Node
\draw (197.6,117.73) node [anchor=north west][inner sep=0.75pt]  [font=\Large]  {$\cdots $};
% Text Node
\draw (136.51,193.2) node [anchor=north west][inner sep=0.75pt]    {$\mathbf{e}_{1}$};
% Text Node
\draw (172.51,193.2) node [anchor=north west][inner sep=0.75pt]    {$\mathbf{e}_{2}$};
% Text Node
\draw (274.51,193.2) node [anchor=north west][inner sep=0.75pt]    {$\mathbf{e}_{5}$};
% Text Node
\draw (129.8,163) node [anchor=north west][inner sep=0.75pt]    {$g$};
% Text Node
\draw (164.2,163) node [anchor=north west][inner sep=0.75pt]    {$g$};
% Text Node
\draw (267.6,163) node [anchor=north west][inner sep=0.75pt]    {$g$};

\end{tikzpicture}

    \caption{Structure of the Deep Smoother module. A sequence of measurements from a track is processed into a state trajectory $\hat{\mathbf x}_{1:T}$, existence probabilities $p_{1:T}$, and a global trajectory existence probability $\bar{p}$.}
    \label{fig:deep_smoother}
    \vspace{-10pt}
\end{figure}
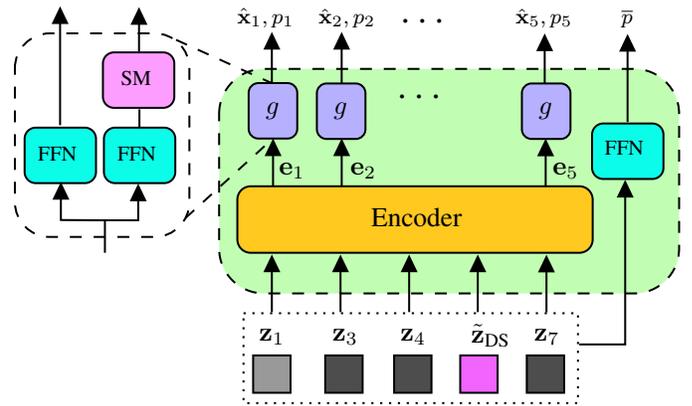
The DS module is in charge of processing each track of measurements into the parameters of the output multi-Bernoulli density of trajectories, i.e., $\{ \hat{\mathbf x}_{1:T}^i, p_{1:T}^i, \bar{p}^i \}_{i=1}^B$. To achieve this, we use a transformer encoder for processing a track's measurements, followed by nonlinearities applied to each time-step. Fig.\,\ref{fig:deep_smoother} illustrates the application of the DS module to the track shown in the top-right of Fig.\,\ref{fig:overview}. The encoder generates an embedding sequence $\mathbf e_{1:T}$ ($T=5$ in the example), and each embedding is fed to a separate head for predicting the state $\hat{\mathbf x}_t$ and the auxiliary existence variable $p_t$ at each time-step $t$. The encoder uses its self-attention to efficiently attend to all measurements in a track when predicting $(\hat{\mathbf x}_t, p_t)$, $t\in\mathbb N_T$. To deal with time-steps with missing measurements, we draw inspiration from Masked Language Modeling \cite{BERT} and feed a dummy vector $\tilde{\mathbf z}_\text{DS}$ to all such time-steps. Similar to the DDA module, the measurements (with confidences) are projected into a higher-dimensional representation with the help of an FFN before feeding them to the encoder (not shown in the figure to avoid clutter).

Besides predicting $\hat{\mathbf x}_{1:T}$ and $p_{1:T}$, the DS module also outputs the existence probability $\bar{p}$ for the track's estimated trajectory. This is accomplished by concatenating all the track's measurements (along with confidences) into a high-dimensional vector and feeding it to an FFN, as shown in Fig.\,\ref{fig:deep_smoother}. This allows the smoother to, for instance, decide if a track contains only false measurements and should be discarded.

\vspace{-0.2cm}
\subsection{Losses}
\label{subsec:losses}
This section describes the two losses used to train D3AS, one for specifically supervising the data association output and one for the parameters of the multi-Bernoulli density of trajectories.

\subsubsection{Deep Data Associator Loss}
\label{subsec:dda_loss}

%- In contrast to traditional approaches, this requires neither specification of pairwise costs (kinematic or based on other properties of the states of the objects) for building a cost matrix nor solving (multidimensional) linear assignment problems for finding the optimal data association during inference.
%- mention that we solve for globally optimal assignment
%- mention that we don't need pairwise costs

In order to train the DDA module, we need a loss $\mathcal L_\text{DDA}$ that supervises the predicted association matrix $\mathbf A$ using the measurements $\mathbf z_{1:n}$ and the ground-truth information of which object each measurement came from. This loss should penalize associations where measurements from the same object are put into different tracks, as well as associations where measurements from different objects are put into the same track. At the same time, it should be invariant to the order of the columns in the association matrix $\mathbf A$, since such order is arbitrary from the point of view of the final predicted data associations. Lastly, it is also necessary that the loss is differentiable with respect to the model's parameters, and desirable that it is easy and stable to optimize with algorithms like stochastic gradient descent.

During training, we assume we know which measurements originate from the same object. Based on this knowledge, we can formulate a ground-truth association matrix $\mathbf A^*$ where each column corresponds to a unique object and each row is a one-hot vector.  Measurements from the same object have 1's in the same column, and false measurements are treated as all coming from the same object (different from all the other objects). We can then use cross-entropy as a loss function for training $\mathbf A$: 
\begin{equation}
    \mathcal L_\text{DDA}(\mathbf A, \mathbf A^*) = -\frac{1}{N}\sum_{i=1}^N    \sum_{j=1}^B\log(\mathbf A_{i,j})\cdot \mathbf A^*_{i,j}~.
    \label{eq:dda_loss}
\end{equation}
To obtain a loss function that is invariant to how we order the columns in $\mathbf A$, we can select $\mathbf A^*$ by ordering its columns to match the tracks predicted by $\mathbf A$. This can be done by solving a minimum cost assignment between tracks and objects, where the cost of matching a track to an object is related to how much mass of measurements from that object the matrix $\mathbf A$ has put on that track. 

Concretely, we first solve the following assignment between $B$ tracks and $|\mathbb T_T|$ objects:
\begin{align}
    \mathbf S^* = \arg\min_{\mathbf S} &\sum_{j=1}^B\sum_{k=1}^{|\mathbb T_T|} \mathbf S_{j,k} \cdot \mathbf C_{j,k}~,
    \label{eq:s_star_minimum_cost_match}
\end{align}
where $\mathbf S$ is an assignment matrix subject to the constraints
\begin{align}
    \sum_{j=1}^B\mathbf S_{j,k}=1,
    \quad
    \mathbf S_{j,k}\in\{0,1\}~,
\end{align}
for all $j\in\mathbb N_B$, and all $k\in\mathbb N_{|\mathbb T_T|}$. The cost matrix $\mathbf C$ is defined as
\begin{equation}
    \mathbf C_{j,k} = -\sum_{i=1}^n \mathbf A_{i, j}\cdot \mathbb I(b_i=k)~,
\end{equation}
where $\mathbb I(\cdot)$ is an indicator function taking the value 1 if the argument is true, 0 otherwise. The symbol $b_i$ denotes the ground-truth object index for measurement $i$ (false measurements are all given the ground-truth object index $-1$). In essence, this linear program attempts to associate object $j$ to the track $i$ that contains most of the mass of the measurements coming from object $j$ (according to $\mathbf A$). Then, given $\mathbf S^*$, we compute $\mathbf A^*_{i,j}=\mathbb I(b_i=s^*_j)$,
%\begin{equation}
%    \mathbf A^*_{i,j}=\mathbb I(b_i=s^*_j)
%\end{equation}
where
\begin{equation}
    s^*_j=\left\{
    \begin{matrix}
        \arg\underset{k}{\max}~\mathbf S^*_{j,k}~, & \text{if } \underset{k}{\max}~\mathbf S_{j,k}=1
        \\
        -1~, & \text{otherwise.} 
    \end{matrix}\right.
    \label{eq:s_star}
\end{equation}

%We note that before obtaining the formulation of $\mathcal L_\text{DDA}$ proposed above, we created 11 different versions of it\footnote{The interested reader is referred to the GitHub page for this project, where we make all of the variants tested available for further exploration.} before finally obtaining a loss that satisfies all the criteria above. These prior attempts suffered from multiple types of problems during optimization, such as diverging gradient steps, attractive local optima, improper handling of uncertainties in the final model, over-sensitivity to initial conditions, over-sensitivity to the number of tracks, etc., rendering them unreliable for training. The version used in this paper was the only, among the ones tested, that led to stable training and high-performing models under a variety of starting conditions and hyperparameters.

\subsubsection{Deep Smoother Loss}
We train the DS module on a loss $\mathcal L_\text{DS}$, defined as the negative log-likelihood of the ground-truth $\mathbb T_T$, given the predicted multi-Bernoulli RFS density parameters $\{\hat{\mathbf x}_{1:T}^i, p_{1:T}^i, \bar p^i\}_{i=1}^B$, as proposed in \cite{nll_letter}. To maintain computational tractability, we approximate the NLL using only the most likely association between predictions and ground-truth, which in turn we approximate as the match given by $\mathbf S^*$ computed in \eqref{eq:s_star}. The loss becomes:
\begin{align}
    &\mathcal L_\text{DS}\Big(\{\hat{\mathbf x}_{1:T}^i, p_{1:T}^i, \bar p^i\}_{i=1}^B, \mathbb T_T \Big) = \nonumber
    \\
    \sum_{i=1}^B &\left\{
    \begin{matrix}
        -\log(1-\bar{p}^i) & \text{if }\mathbb T_T^i=\varnothing
        \\
        -\log(\bar{p}^i) + 
        \mathcal L_\text{nll}(\hat{\mathbf x}_{1:T}^i, p_{1:T}^i,\mathbb T_T^i) & \text{otherwise,}
    \end{matrix}
    \right.
    \label{eq:smoother_loss}
\end{align}
where $\mathbb T_T^i$ denotes the ground-truth trajectory $(t_s^i, \mathbf x_{t_s^i+l^i})$ in $\mathbb T_T$ matched to prediction $i$, according to $\mathbf S^*$. If no ground-truth trajectories are matched to prediction $i$, we define $\mathbb T_T^i=\varnothing$. The term $\mathcal L_\text{nll}(\hat{\mathbf x}_{1:T}^i, p_{1:T}^i, \mathbb T_T^i)$ denotes the NLL of trajectory $\mathbb T_T^i$ given the parameters of the $i$-th prediction. To compute this, we interpret the predictions $(\hat{\mathbf x}_t^i, p_t^i)$ at each time-step $t$ as independent Bernoulli RFSs, each with existence probability $p_t^i$ and state density $\mathcal N(\hat{\mathbf x}_t^i, \mathbf I)$, where $\mathbf I$ denotes an identity matrix. Hence, we have that $\mathcal L_\text{nll}$ becomes:
\begin{equation}
    \mathcal L_\text{nll}(\hat{\mathbf x}_{1:T}^i, p_{1:T}^i, \mathbb T_T^i) = \sum_{t=1}^T f(\hat{\mathbf x}_t^i, p_t^i, \mathbb T_T^i)~,
\end{equation}
where 
\begin{equation}
    f(\hat{\mathbf x}_t^i, p_t^i, \mathbb T_T^i) = 
    \left\{
    \begin{matrix}
        \Vert\hat{\mathbf x}_{t}^i-\mathbf x_{t}^i\Vert_2^2 -\log(p_{t}^i) & \text{if }t_s^i\leq t \leq t_s^i+l^i
    \\
        -\log(1-p_t^i) & \text{otherwise}.
    \end{matrix}
    \right.
\end{equation}

%In essence, this loss trains the filter to correctly predict the states of the objects for which each of its tracks was matched to. We note that for these types of models, it is customary to find the match between predicted tracks and objects by comparing the state trajectories $(\hat{\mathbf x}_{1:T}, p_{1:T})$ with each of the trajectories, and matching each to its closest one (e.g., in L2 sense). That is, the matching object for a certain track is determined based on the filter's output $(\hat{\mathbf x}_{1:T}, p_{1:T})$ for that track. This means that as the DS trains, the matches for its predictions (which are used as targets for the predictions) are also changing, which has been identified as a potential cause for high-variance loss signals (cite the denoising papers here), impacting the training. We, on the other hand, use the filter inputs (the data associations for each track) to determine the correct match for that track, therefore guaranteeing that it stays fixed regardless of the DS's parameters changing.

%\subsection{Preprocessing}
%\label{subsec:preprocessing}
%As usual in deep learning applications, we normalize the measurements so that all dimensions vary across approximately the same range. We do so by using the size of the field of view and the birth model covariances as estimates for the usual range of the different measurement dimensions. 

\section{EVALUATION SETTING}
\label{sec:evaluation_setting}
This section describes the evaluation protocol used, starting with a description of the tasks in which the evaluation takes place, followed by implementation details of each of the benchmarked algorithms, and closing with a mathematical description of the performance metrics used.

\subsection{Task Description}
\label{subsec:task_description}
In order to compare the performance of D3AS to the model-based Bayesian benchmark under a variety of different conditions, we create ten different tasks for the evaluations. All tasks have the same multi-object models, but certain parameters are changed to vary their challenge levels. 

The motion model used in all tasks is the nearly constant velocity model, defined as:
\begin{equation}
    f(\mathbf x^{t+1}|\mathbf x^t) = 
    \mathcal N\left(\mathbf x^{t+1}; \begin{bmatrix}
        \mathbf I & \mathbf I\Delta_t
        \\ 
        \mathbf 0 & \mathbf I
    \end{bmatrix} \mathbf x^t~,~\sigma_q^2
    \begin{bmatrix}
        \mathbf I\frac{\Delta_t^3}{3} & \mathbf I\frac{\Delta_t^2}{2}
        \\
        \mathbf I\frac{\Delta_t^2}{2} & \mathbf I\Delta_t
    \end{bmatrix}\right),
\end{equation}
where $\mathbf x^t, \in \mathbb R^{d_x}$, $d_x=4$ represents object position and velocity in 2D at time-step $t$, $\Delta_t=0.1$ is the sampling period, and $\sigma_q$ controls the magnitude of the process noise. New objects are sampled according to a Gaussian birth model which is a PPP, with Poisson rate $\bar{\lambda}_0=6$ and spatial distribution $\mathcal N(\boldsymbol\mu_b, \boldsymbol\Sigma_b)$,
\begin{equation*}
    \boldsymbol\mu_b=\begin{bmatrix}
        7\\ 
        0\\ 
        0\\ 
        0
    \end{bmatrix},
    \quad 
    \boldsymbol\Sigma_b=\begin{bmatrix}
        10 & 0 & 0 & 0\\ 
        0 & 30 & 0 & 0\\ 
        0 & 0 & 16 & 0\\ 
        0 & 0 & 0 & 16
    \end{bmatrix},
\end{equation*}
with values chosen to have an object birth model that covers most of the field-of-view. The measurement model used is a non-linear Gaussian model simulating a radar system: $g(\mathbf z | \mathbf x)=\mathcal N\big(\mathbf z; H(\mathbf x), \boldsymbol\Sigma(\mathbf x)\big),$ where $H$ transforms the 2D-position and velocity vector $\mathbf x$ into $(r, \dot{r}, \theta)$, respectively the range, Doppler and bearing of each object. The covariance $\boldsymbol\Sigma(\mathbf x)$ is state-dependent and modeled to provide less accurate measurements closer to the edges of the FOV, similar to the measurement noise in radar applications. To do so, we use quadratic functions on $r$ and $\theta$:
\begin{align}
    \boldsymbol\Sigma(\mathbf x) &=
    \begin{bmatrix}
        f_1(r) & 0 & 0 \\
        0 & f_2(\dot{r}) & 0 \\
        0 & 0 & f_3(\theta)
    \end{bmatrix}
    \nonumber
    \\
    f_1(r) &= \frac{\sigma^2_{r,\text{max}}-\sigma^2_{r,\text{min}}}{14.5^2}(r-0.5)^2+\sigma^2_{r,\text{min}}
    \nonumber
    \\
    f_2(\dot{r}) &= \sigma^2_{\dot{r}}
    \nonumber
    \\
    f_3(\theta) &= \frac{\sigma^2_{\theta, \text{max}}-\sigma^2_{\theta, \text{min}}}{1.3^2}\theta^2 + \sigma^2_{\theta, \text{min}}
    \label{eq:meas_noise_cov}~,
\end{align}
where $(r, \dot{r}, \theta)=H(\mathbf x)$, and the coefficients of these quadratic forms were chosen to make $\boldsymbol\Sigma$ have the value $\text{Diag}(\sigma^2_{r,\text{min}}, \sigma^2_{\dot{r}}, \sigma^2_{\theta, \text{min}})$ close to the sensor, and $\text{Diag}(\sigma^2_{r,\text{max}}, \sigma^2_{\dot{r}}, \sigma^2_{\theta, \text{max}})$ at the edges of the FOV. All tasks have $\sigma^2_{r, max}=\sigma^2_{\theta, max}=0.04$.

Lastly, all tasks have a measurement window of $T=10$ time-steps, and an FOV delimited by the intervals $(0.5, 15), (0, 8), (-1.3, 1.3)$ for $r$ (m), $\dot{r}$ (m/s), and $\theta$ (radians), respectively. The probability of survival is constant inside the FOV with value $p_s(\mathbf x)=0.98$. Both the probability of survival $p_s$ and the detection probability $p_d$ are zero outside the FOV.

The task-specific hyperparameters used to change the challenge level of the tasks are shown in Table \ref{tab:data_association_hyperparam}. Tasks 1-4 all have relatively high probability of detection and become increasingly difficult in terms of the clutter rate $\lambda_c$ and their probability of detection $p_d$. Tasks 5-10 have lower $p_d$ and considerably higher process and/or measurement noise, depending on the tasks.

\begin{table}[h]
\centering
\caption{Task-specific hyperparameters.
\label{tab:data_association_hyperparam}}
\begin{tabular}{@{}ccccccc@{}}
    \toprule
    \textbf{Task} & 
    $p_d$ & 
    $\sigma_q^2$ & 
    $\sigma^2_{r,\text{min}}$ & 
    $\sigma^2_{\dot{r}}$ & 
    $\sigma^2_{\theta,\text{min}}$ & 
    $\lambda_c$
    \\ 
    \midrule
    1   & $0.99$  &  $1.0$  &  $10^{-4}$  &  $0.01$   &  $10^{-4}$  &  $1.6\cdot10^{-2}$  \\
    2   & $0.99$  &  $1.0$  &  $10^{-4}$  &  $0.01$   &  $10^{-4}$  &  $3.2\cdot10^{-2}$  \\
    3   & $0.85$  &  $1.0$  &  $10^{-4}$  &  $0.01$   &  $10^{-4}$  &  $6.6\cdot10^{-2}$  \\
    4   & $0.99$  &  $1.0$  &  $10^{-4}$  &  $0.01$   &  $10^{-4}$  &  $1.3\cdot10^{-1}$  \\
    5   & $0.70$  &  $4.0$  &  $10^{-4}$  &  $0.01$   &  $10^{-4}$  &  $6.6\cdot10^{-2}$  \\
    6   & $0.70$  &  $2.0$  &  $10^{-3}$  &  $1.00 $  &  $10^{-3}$  &  $6.6\cdot10^{-2}$  \\
    7   & $0.60$  &  $3.0$  &  $10^{-2}$  &  $1.00 $  &  $10^{-2}$  &  $6.6\cdot10^{-2}$  \\
    8   & $0.70$  &  $3.0$  &  $10^{-4}$  &  $0.01$   &  $10^{-4}$  &  $1.3\cdot10^{-1}$  \\
    9   & $0.70$  &  $1.0$  &  $10^{-3}$  &  $1.00 $  &  $10^{-3}$  &  $1.3\cdot10^{-1}$  \\
    10  & $0.60$  &  $3.0$  &  $10^{-2}$  &  $1.00 $  &  $10^{-2}$  &  $1.3\cdot10^{-1}$  \\
    \bottomrule
\end{tabular}
\vspace{-10pt}
\end{table}

\subsection{Implementation Details}
\label{subsec:implementation_details}
This section describes the implementation details for the deep-learning smoother D3AS and the model-based Bayesian benchmark TPMBM. 

\subsubsection{D3AS}
\label{subsec:D3AS_implementation_details}
In all the experiments, we increase the dimensionality of the measurements to $128$ for both modules using a linear projection and use $6$ encoder layers with $8$ attention heads. The FFNs in the encoders of both modules have 2048 hidden units and have a dropout rate of $0.1$. In the DDA module, the final FFN layer to project embeddings into measurement pmfs has 3 layers with 128 hidden units each, written as 3$\times$128 as a shorthand, and we use $B=20$ as the number of maximum possible tracks. In the DS module, we use two separate FFNs of size 3x128 each for predicting the position and velocity from the embeddings, one 2x64 for the existence probabilities $p_{1:T}$, and another 2x64 for the trajectory existence probability $\bar{p}$.

Instead of feeding the measurements $\mathbf z_i$ from each partition into the decoder, we first process each into $\tilde{\mathbf z}_i$, where
\begin{equation}
    \tilde{\mathbf z}_i = (r\cos\theta, r\sin\theta, \dot{r})~,
\end{equation}
i.e., using the range and bearing dimensions of the measurements to compute the corresponding 2D Cartesian position and leaving the doppler component $\dot{r}$ untouched. Empirical results show that this accelerates learning (with negligible impact to final performance) because it frees the DS module from the additional burden of learning the nonlinear mapping from the measurement space to the state space.

The training of D3AS is done by first training the DDA module for $2\cdot10^6$ gradient steps with a batch size of $32$. Then, the DS module uses the trained DDA's associations and is trained for $2\cdot10^6$ gradient steps (on the same samples), with a batch size of $16$. Both training procedures use AdamW \cite{adamw} with a learning rate of $5\cdot10^{-5}$. If the moving average of the loss signal (window of $2\cdot10^3$ gradient steps) does not decrease for $10^5$ consecutive gradient steps, the learning rate is automatically reduced by a factor of 2. To create a set of trajectories from D3AS's output, we only keep trajectories with $\bar{p}>0.5$, and only keep time-steps with $p_t>0.8$. The code is implemented in Python/Pytorch, and training is performed on a V100 GPU, taking approximately two days to complete for each module. All the code required to define, train, and evaluate the models is made publicly available at \url{https://github.com/JulianoLagana/TBD}.

\subsubsection{TPMBM}
%- describe TPMBM implementation details here, including how we extract trajectories to later compute TGOSPA.
We consider the TPMBM implementation for the set of all trajectories \cite{granstrom2018poisson}, which produces full trajectory estimates upon receipt of each new set of measurements. 
%TPMBM uses a Poisson birth model with Poisson intensity $\lambda_b\mathcal{N}(\boldsymbol{\mu}_b,\boldsymbol{\Sigma}_b)$, and the initial Poisson intensity for undetected objects is set to $\lambda_0\mathcal{N}(\boldsymbol{\mu}_b,\boldsymbol{\Sigma}_b)$. 
To achieve computational tractability of TPMBM, it is necessary to reduce the number of parameters used to describe the posterior densities. First, gating is used to remove unlikely measurement-to-object associations, by thresholding the squared Mahalanobis distance, where the gating size is 20. Second, we use Murty’s algorithm \cite{crouse2016implementing} to find up to 200 best global hypotheses and prune hypotheses with weight smaller than $10^{-4}$. Third, we prune Bernoulli components with probability of existence smaller than $10^{-5}$ and Gaussian components in the Poisson intensity for undetected objects with weight smaller than $10^{-5}$. Moreover, we do not update Bernoulli components with probability smaller than $10^{-4}$ of being alive at the current time-step.

TPMBM performs smoothing-while-filtering without $L$-scan approximation \cite{garcia2020trajectory}. To handle the nonlinearity of the measurement model, the iterated posterior linearization filter (IPLF) \cite{garcia2015posterior} is incorporated in both PMBM and $\delta$-GLMB, see, e.g., \cite{garcia2021gaussian}. The IPLF is implemented using sigma points with the fifth-order cubature rule \cite{arasaratnam2009cubature} as suggested in \cite{crouse2014basic} for radar tracking with range-bearing-Doppler measurements, and the number of iterations is 5. In IPLF, the state-dependent measurement noise covariance $\Sigma(\mathbf x)$ is approximated as $\Sigma(\hat{\mathbf x})$ where $\hat{\mathbf x}$ is either the mean of the predicted state density or the mean of the state density at last iteration. 

To extract trajectory estimates from the TPMBM posterior, we first select the global hypothesis with the highest weight, and then we report the starting times and means with most likely duration of the Bernoulli components whose probability of existence is greater than $0.5$.
The TPMBM implementation was developed in MATLAB, adapted from the code available at \url{https://github.com/Agarciafernandez/MTT/tree/master/TPMBM\%20filter}.

\subsection{Performance Metrics}
We use two performance metrics for evaluating the algorithms considered. The first one, TGOSPA, evaluates overall trajectory estimation performance, while the second one, top-1 association accuracy, evaluates the performance of the methods only on the data association subtask.

\subsubsection{TGOSPA}
The trajectory estimation performance is evaluated using Trajectory-GOSPA \cite{tgospa}, which is an extension of the GOSPA metric \cite{GOSPA} to sets of trajectories. Its values range from 0 to $\infty$, with lower values corresponding to better tracking performance. The trajectory-GOSPA metric penalizes localization costs for properly detected objects, misdetections, false detections, and track switches. 

Let $\Pi_{\mathbf{X},\mathbf{Y}}$ be the set of all possible assignment vectors between the index sets $\{1.\dots,|\mathbf{X}|\}$ and $\{0,\dots,|\mathbf{Y}|\}$. An assignment vector $\pi^k = [\pi_1^k,\dots,\pi^k_{|\mathbf{X}|}]^{\text{T}}$ at time-step $k$ is a vector $\pi^k\in\{0,\dots,|\mathbf{Y}|\}^{n_{\mathbf{X}}}$ such that its $i$th component $\pi_i^k=\pi^k_{i^\prime}=j>0$ implies that $i=i^\prime$. Here $\pi_i^k=j\neq 0$ implies that trajectory $i$ in $\mathbf{X}$ is assigned to trajectory $j$ in $\mathbf{Y}$ at time-step $k$ and $\pi_i^k=0$ implies that trajectory $i$ in $\mathbf{X}$ is unassigned at time-step $k$.

For $1\leq p<\infty$, cut-off parameter $c > 0$, switching penalty $\gamma > 0$ and a base metric $d_b(\cdot,\cdot)$ in the single object space $\mathfrak{X}$, the multidimensional assignment metric $d_p^{(c,\gamma)}(\mathbf{X},\mathbf{Y})$ between two sets $\mathbf{X}$ and $\mathbf{Y}$ of trajectories in time interval $1,\dots,T$ is \vspace{-10pt}
\begin{multline}
    \label{eq_tra_metric_mda}
    d_{p}^{(c, \gamma)}(\mathbf{X}, \mathbf{Y}) =
    \\ 
    \min _{\pi^{k} \in \Pi_{\mathbf{X}, \mathbf{Y}}}
    \Big(
        \sum_{k=1}^{T} d_{\mathbf{X}, \mathbf{Y}}^{k}(\mathbf{X}, \mathbf{Y}, \pi^{k})^{p} + 
        \sum^{T-1}_{k=1} s_{\mathbf{X},\mathbf{Y}} (\pi^k,\pi^{k+1})^p 
    \Big)^{\frac{1}{p}}~,
\end{multline}
where the costs (to the $p$-th power) for properly detected objects, misdetections, and false detections at time-step $k$ are 
\begin{align}\label{eq_tra_metric_loc}
    &d_{\mathbf{X}, \mathbf{Y}}^{k}\left(\mathbf{X}, \mathbf{Y}, \pi^{k}\right)^{p}=\sum_{(i, j) \in \theta^{k}\left(\pi^{k}\right)} d\left(\tau^k(X_i), \tau^k(Y_j)\right)^{p} \nonumber
    \\
    &+\frac{c^{p}}{2}\left(\left|\tau^{k}(\mathbf{X})\right|+\left|\tau^{k}(\mathbf{Y})\right|-2\left|\theta^{k}\left(\pi^{k}\right)\right|\right)~,
\end{align}
with 
\begin{multline}
    \theta^{k}\left(\pi^{k}\right)=
    \left\{
        \left(i, \pi_{i}^{k}\right): i \in\left\{1, \ldots, n_{\mathbf{X}}\right\}, \right. \\\left. |\tau^k(X_i)| = \big|\tau^k(Y_{\pi^k_i})\big| = 1, d\big( \tau^k(X_i),\tau^k(Y_{\pi^k_i})\big) < c  
    \right\}
\end{multline}
and the switching cost (to the $p$-th power) from time-step $k$ to $k+1$ is given by 
\begin{align}
    &s_{\mathbf{X},\mathbf{Y}}(\pi^k,\pi^{k+1})^p = \gamma^p\sum^{|\mathbf{X}|}_{i=1}s\left( \pi_i^k,\pi_i^{k+1} \right)
    \\
    &s\left( \pi_i^k,\pi_i^{k+1} \right) = \begin{cases}
        0 & \text{if } \pi_i^k = \pi_i^{k+1}\\
        1 & \text{if } \pi_i^k \neq \pi_i^{k+1}, \pi_i^k\neq 0, \pi_i^{k+1} \neq 0\\
        \frac{1}{2} & \text{otherwise.}
    \end{cases}.
\end{align}
It should be noted that for $(i,j)\in\theta^k$, $\tau^k(X_i)$ and $\tau^k(Y_{\pi^k_i})$ contain precisely one element, and their distance is smaller than $c$, so $d\left(\tau^k(X_i), \tau^k(Y_j)\right)$ coincides with $d_b(\cdot,\cdot)$ evaluated at the corresponding single object states, which corresponds to the localization error. Therefore, (\ref{eq_tra_metric_loc}) represents the sum of the costs (to the $p$th power) that correspond to localization error for properly detected objects (indicated by the assignments in $\theta^k(\pi^k)$), number of misdetections $(|\tau^k(\mathbf{X})|-|\theta^k(\pi^k)|)$ and false detections $(|\tau^k(\mathbf{Y})|-|\theta^k(\pi^k)|)$ at time-step $k$. Trajectory-GOSPA metric is implemented using linear programming, and the hyperparameters used in the experiments are: $p=1$, cut-off parameter $c=20$, switching penalty $\gamma > 0$, and $L1$ norm as the base metric in the single object space.

\subsubsection{Top-1 Association Accuracy}
\label{subsec:taa}
In order to separately evaluate the performance of each method on the data association task, we define the \textit{top-1 association accuracy} (TAA) performance measure on a data association matrix $\mathbf A\in\mathbb R^{N\times B}$ associating $N$ measurements to $B$ tracks. 

To do so, we first compute the optimal association between tracks and measurements according to the ground-truth associations between measurements and objects. This is done exactly as described in Sec.\,\ref{subsec:dda_loss}, i.e., solving the minimum assignment problem defined in \eqref{eq:s_star_minimum_cost_match}. Let $s_j\in\mathbb N$ be the object index to which track $j$ is matched, and $b_i\in\mathbb Z$ the index of the object that originated measurement $i$ ($-1$ for false measurements). Then, TAA is defined as:
\begin{align}
    \text{TAA}(\mathbf A) &= \frac{\sum_{i=1}^N \mathbb I(s_{j^*(i)} = b_i) \mathbb I(b_i\neq -1) }{\sum_{i=1}^N \mathbb I(b_i\neq -1)} 
    \\
    j^*(i) &= \arg\max_j \mathbf A_{i, j}~.
\end{align}
That is, for each measurement $i$, we find the top track it is associated to ($j^*(i)$) and compare the object matched to that track ($s_{j^*(i)}$) with the ground-truth object for measurement $i$ ($b_i$). The terms $\mathbb I(b_i\neq -1)$ are added to ignore the class of false measurements when computing accuracy. This is done because, in most tasks, the number of false measurements is considerably larger than the number of true measurements, skewing the accuracy results upwards and making them less informative. TAA ranges from 0 to 1, corresponding to a data association in which all measurements were associated incorrectly or correctly, respectively.

\section{RESULTS}
\label{sec:results} 
This section describes the experimental results obtained, comparing the deep-learning smoother with the model-based Bayesian counterpart. 

\subsection{Training the DDA module}
\begin{figure}[t]
    \centering
    \includegraphics[width=0.48\textwidth]{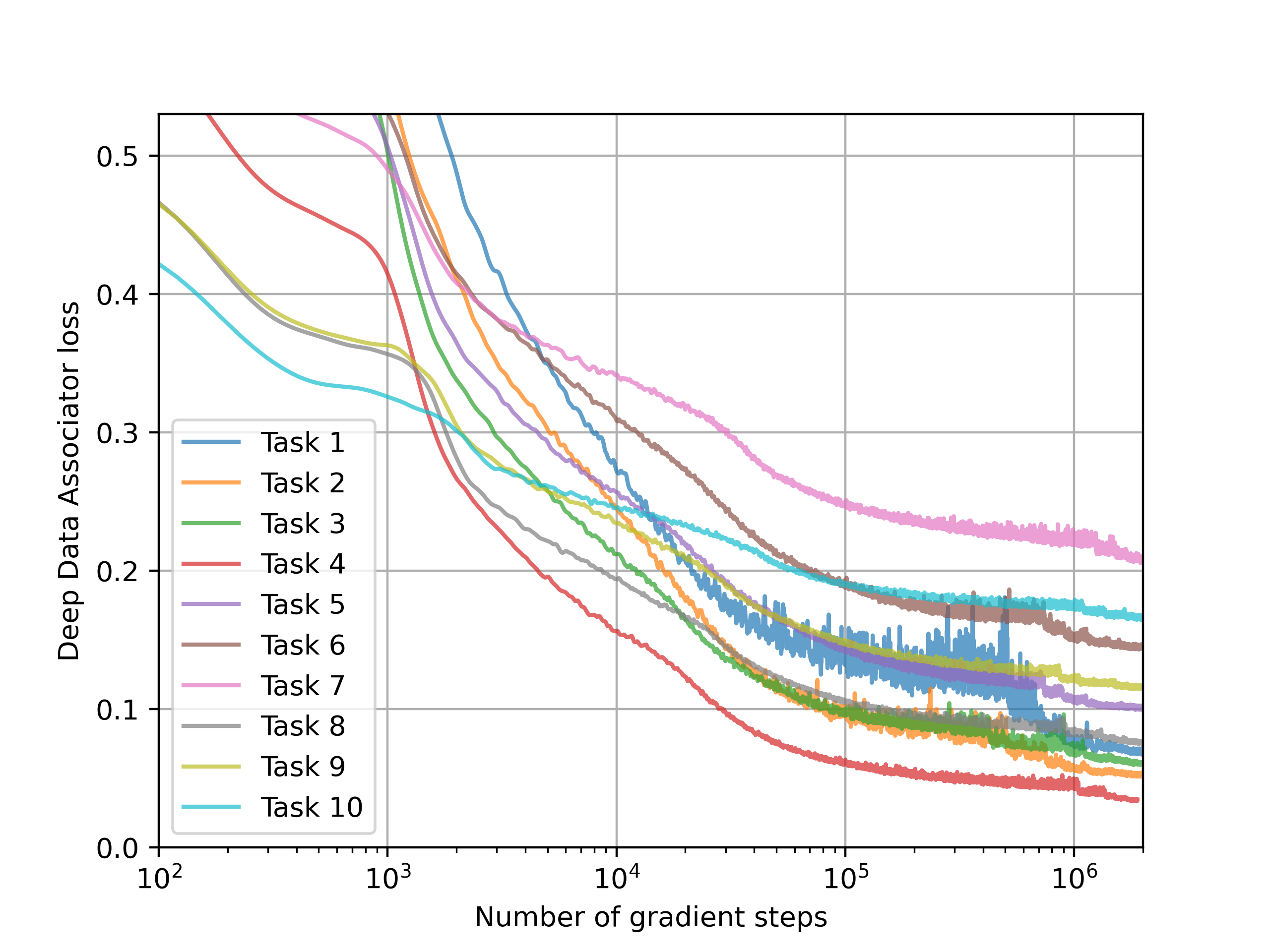}
    \caption{Data association loss $\mathcal L_\text{DDA}$ during training of the DDA module. Lower values indicate better data association performance.}
    \label{fig:dda_losses}
    \vspace{-10pt}
\end{figure}

The training curves for the DDA module in each task are shown in Fig.\,\ref{fig:dda_losses}. Here we plot the data association loss $\mathcal L_\text{DDA}$, as defined in \eqref{eq:dda_loss}, versus the the number of gradient steps taken (in logarithmic scale). Loss curves have been smoothed with a Gaussian kernel ($\sigma=100$) to ease interpretation.
\begin{figure}[t]
    \centering
    \includegraphics[width=0.48\textwidth]{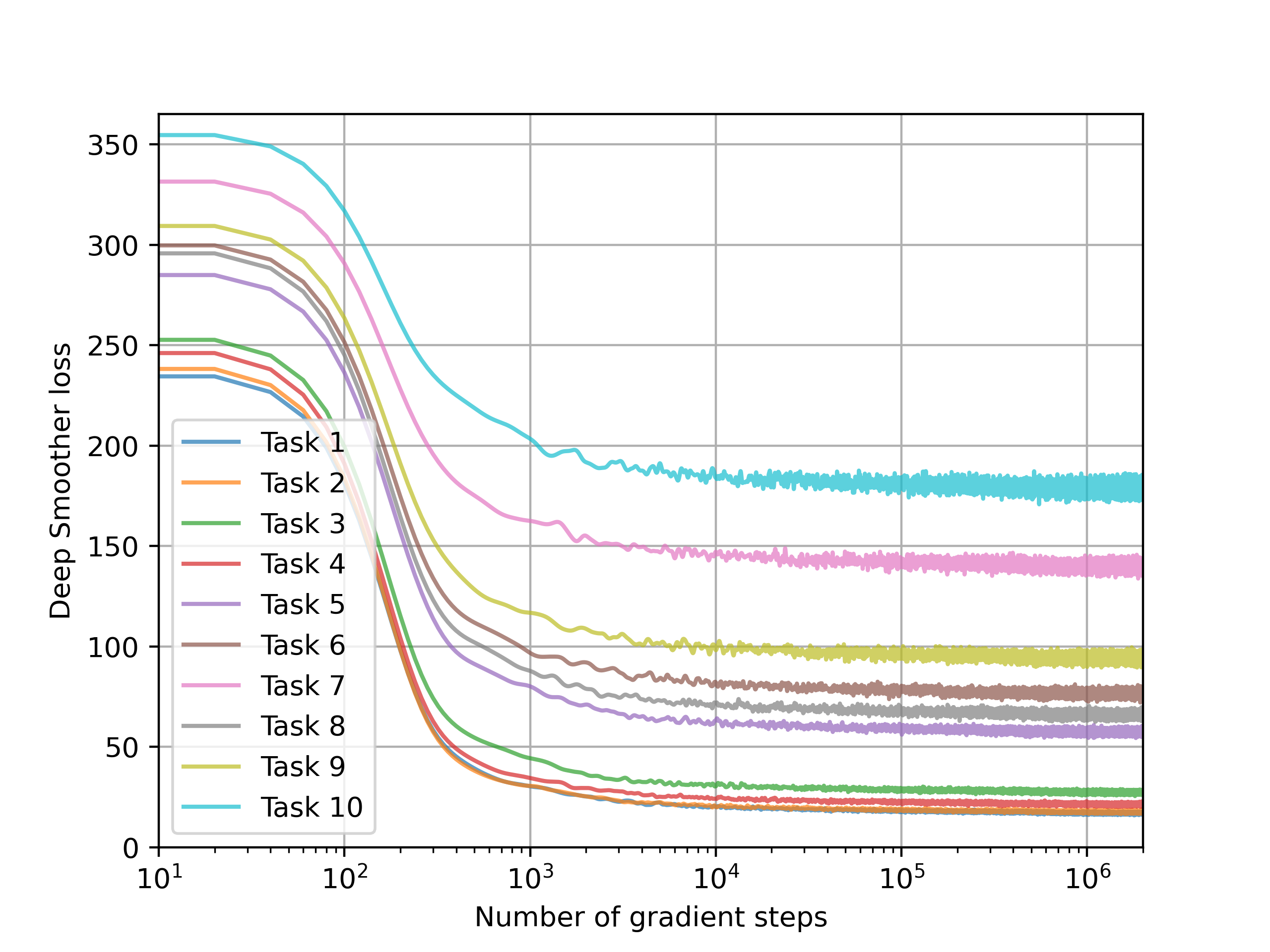}
    \caption{Deep Smoother loss $\mathcal L_\text{DS}$ values for D3AS in each task vs. number of gradient steps that the DS module was trained for. Lower values indicate better trajectory estimates.}
    \label{fig:df_losses}
    \vspace{-10pt}
\end{figure}

The loss $\mathcal L_\text{DDA}$ drops sharply at the beginning of the training and starts to slow down for most of the tasks around $10^{5}$ gradient steps. The rate of decrease varies during the optimization, which we believe is due to the model learning different aspects of the task at different points in time. Visual inspection of the model's output during training showed that it first learns to put almost all the measurements' mass in the same track. This is due to the fact that, in most tasks, the most prevalent class are false measurements, which makes this a simple way to reduce the loss from a random initialization of the model's weights. Importantly, the model does not get stuck in this initial state, and a bit later in the training, the DDA slowly starts to try separating true measurements from this track into different tracks, first with easier examples (trajectories with more measurements or in parts of the FOV with fewer false measurements) and later with harder ones.

The sudden decreases close to $10^6$ gradient steps are due to the automatic learning rate reduction (see Sec.\,\ref{subsec:D3AS_implementation_details}). This reduction in learning rate is not essential when training the DDA, but in our experiments it improved final performance for all tasks. Furthermore, judging from the shape of the loss curves close to the end of the x-axis, we expect the DDA module to benefit from further training in all tasks, but all trials were restricted to $2\cdot10^6$ gradient steps due to computational limits. We also note that the variance of $\mathcal L_\text{DDA}$ varies considerably depending on the task. This is most likely due to the different number of measurements in each task. We can interpret the loss defined in \eqref{eq:dda_loss} as the Monte-Carlo estimate of the expected cross-entropy between $\mathbf A$ and $\mathbf A^*$, where increasing $N$ decreases the variance of this estimate.

Lastly, we observe that $\mathcal L_\text{DDA}$ is not comparable among tasks with different average number of false measurements, as increasing this number makes it easier for the models to achieve a lower cross-entropy, since most of the measurements will be from the same false measurement class. However, comparing tasks with the same clutter rate -- i.e., tasks 3, 5, 6, and 7, or tasks 8, 9, and 10 -- shows that indeed increasing measurement or process noise, or decreasing the detection probability makes training harder, and results in worse final data association performance, as expected.

\subsection{Training the DS module}
Once the DDA module is fully trained, its weights are kept fixed, and then the DS module is trained. The loss curves for this procedure are shown in Fig.\,\ref{fig:df_losses}, where we plot the smoothing loss $\mathcal L_\text{DS}$, as defined in \eqref{eq:smoother_loss}, versus the number of gradient steps taken (in logarithm scale). In contrast to $\mathcal L_\text{DDA}$, most gains in $\mathcal L_\text{DS}$ come in the first $10^4$ gradient steps, and training beyond $2\cdot10^6$ seems unlikely to yield any additional performance. This, together with the simple sigmoid shape of the loss curves, suggests that optimizing for $\mathcal L_\text{DS}$ is considerably easier than for $\mathcal L_\text{DDA}$, which agrees with the intuition that once the data association sub-task is solved, smoothing the individual trajectories is easier.

Furthermore, we note that the ordering of the loss curves is different from the ordering in Fig.\,\ref{fig:dda_losses}. The relative challenge in training for each task now depends much less on the clutter rate, and more on the detection probability and measurement/motion noise. For instance, for tasks 1 and 2, which share most of the parameters in Table \ref{tab:data_association_hyperparam} except the clutter rate, which is doubled for task 2, have very similar loss curves (almost overlapping). This indicates that for these tasks, the DDA module is fairly good at removing false measurements from the DS's input, effectively making them the same task from the DS's perspective. On the other hand, more complicated tasks like tasks 7 and 10 have differing loss curves even though they share all parameters but the clutter rate, indicating that the DDA was not able to completely remove the effect of clutter in the DS's input.

\begin{figure*}
    \centering
    \includegraphics[width=0.9\textwidth]{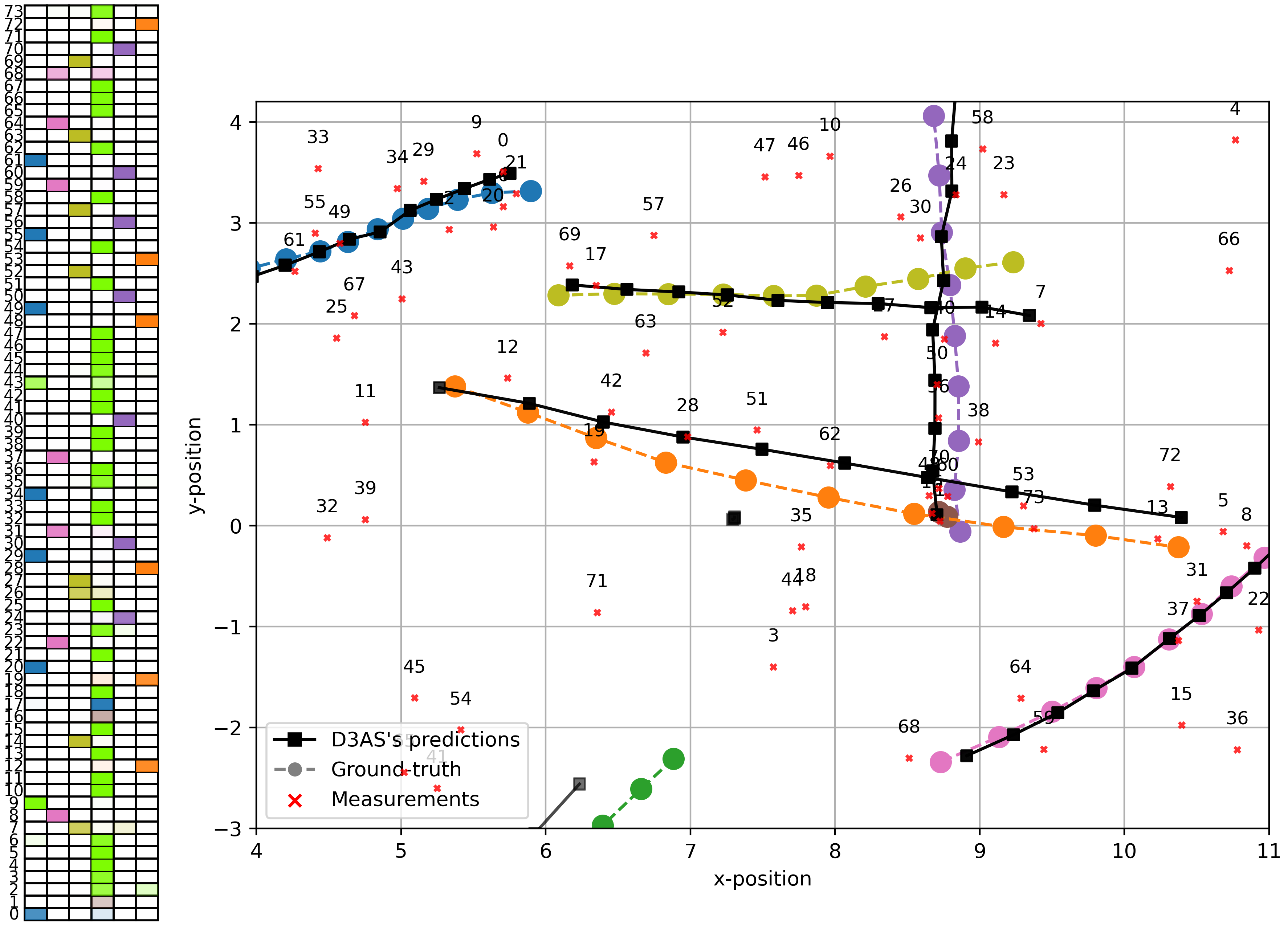}
    \caption{Sample output from D3AS in task 6. Left: data association matrix predicted by the DDA module (tracks with low mass for all measurements are not shown due to limited space). Right: Ground-truth and predicted trajectories by the DS module.}
    \label{fig:D3AS_output}
    \vspace{-10pt}
\end{figure*}

\subsection{Visualization of the trained D3AS tracker}
After both modules are trained, D3AS is able to leverage all the measurements available to predict reasonable trajectories in all the tasks. A sample output for task 6 is illustrated in Fig.\,\ref{fig:D3AS_output}. On the left, the data association matrix predicted by the DDA module is shown, and on the right a visualization of the DS's module output. In the table corresponding to the left, each row corresponds to one measurement, and each column to one track
\footnote{Due to space limitations, measurements not inside the viewing region on the right of Fig.\,\ref{fig:D3AS_output} are not included as rows in the DA matrix. Tracks with almost no mass from any measurements are also not shown.}. 
Each row is color-matched to the object that generated the corresponding measurement, and the transparency in each column depends on how much mass the pmf has in that track. For example, measurement 68 (the 6th row in the table, from top to bottom) has a pmf which has its mode in track number 2 (where the other measurements from this object also have their modes), but also some mass in track number 4 (where most of the false measurements have their modes), signaling that the DDA is uncertain as to whether this is a false measurement. 

On the right, in the DS's output visualization, we show the $xy$ position of all measurements, ground-truth trajectories, and predicted trajectories in the region with $x$ coordinate ranging from 4 to 11 and $y$ from -3 to 4.2 (velocity vectors are not included due to space limitations). All measurements are depicted with a number on top for correspondence with the DA matrix on the left. Ground-truth trajectories have their colors matched to the colors in the DA matrix. 

Fig.\,\ref{fig:D3AS_output} can give us a glimpse of how hard the tasks are. The only input the algorithms have are the measurements (xy shown as red crosses), and the desired output are the ground-truth trajectories shown. For most ground-truth trajectories, its measurement sequence is considerably noisy, and challenging to tell apart from the false measurements. Nevertheless, the DDA module is able to partition the measurements very well, with only a few true measurements being misclassified as false (measurements 16 and 17) and similarly only a few false measurements misclassified as true (measurements 9 and 43). 
%We stress that the DDA module is estimating the globally optimal association for the measurements, taking all the time-steps into account at once. 
The DDA module is also able to inform downstream tasks of uncertain associations, for example measurements that have their mode on a certain track but still considerable mass in others (e.g., measurements 19, 26, 68). This is important, for instance, in the smoothing task, where the DS module leverages this uncertainty to decide how much to trust each measurement.

The figure also shows how well the DS module can predict trajectories, given the associations computed by the DDA. Even though measurement noise is high, and the low probability of detection caused many trajectories to not have measurements in multiple time-steps (e.g., the orange trajectory only generated 6 measurements in its 10 time-steps), the DS module is still able to predict reasonable state estimates. %Importantly, it also correctly estimates the length of these trajectories.

The sample shown in Fig.\,\ref{fig:D3AS_output}, although hand-picked to illustrate interesting behaviors of both modules compactly, is representative of the tracking quality of D3AS in task 6 and most of the other tasks. The next section provides a thorough evaluation over many samples, and a comparison to TPMBM in all tasks using the TGOSPA metric and the TAA performance measure.

\subsection{Comparison to Model-based Bayesian Benchmark}
\label{subsec:comparison_to_tpmbm}
In order to compare D3AS to TPMBM, we compute the average trajectory-GOSPA metric over $10^3$ samples from each task. Both algorithms are evaluated on the same samples for fairness and reduced variance, and all samples are drawn from the multi-object models defined in \ref{sec:evaluation_setting} using the task-specific hyperparameters shown in Table \ref{tab:data_association_hyperparam}. The resulting trajectory-GOSPA scores (lower is better) and decompositions for all ten tasks, along with 95\% confidence intervals for the means, are displayed in Table \ref{tab:tgospa_scores}.  

\begin{table}[]
    \centering
    \caption{Trajectory-GOSPA scores for all tasks.  \label{tab:tgospa_scores}}
    \setlength\extrarowheight{-4pt}
    \begin{tabular}{@{}lllllll@{}}
        \toprule
        \textbf{Task} & \textbf{Alg.} & \textbf{TGOSPA} & \textbf{Loc} & \textbf{Miss} & \textbf{False} & \textbf{Switch}
        \\
        \midrule
        \multirow{2}{*}{1}
            & D3AS & 106.41 ± 3.06 & 91.11 & 13.97 & 1.12 & 0.21 \\
            & TPMBM & 117.82 ± 3.47 & 100.61 & 16.39 & 0.64 & 0.18 \\ 
            \midrule
        \multirow{2}{*}{2}
            & D3AS & 110.54 ± 3.31 & 90.82 & 17.81 & 1.75 & 0.17  \\
            & TPMBM & 124.90 ± 3.64 & 101.02 & 22.72 & 0.91 & 0.25  \\
            \midrule
        \multirow{2}{*}{3}
            & D3AS & 154.02 ± 4.30 & 113.41 & 38.36 & 1.74 & 0.51  \\
            & TPMBM & 171.12 ± 5.05 & 111.89 & 57.23 & 1.16 & 0.84  \\ 
            \midrule
        \multirow{2}{*}{4}
            & D3AS & 122.40 ± 3.26 & 89.67 & 30.14 & 2.38 & 0.21  \\
            & TPMBM & 152.29 ± 4.64 & 93.47 & 57.84 & 0.77 & 0.21  \\
            \midrule
        \multirow{2}{*}{5}
            & D3AS & 232.27 ± 6.28 & 150.33 & 77.90 & 2.71 & 1.34  \\
            & TPMBM & 261.30 $\pm$ 7.63 & 120.66 & 138.98 & 0.30 & 1.40  \\
            \midrule
        \multirow{2}{*}{6}
            & D3AS & 272.17 ± 6.91 & 163.51 & 104.11 & 2.66 & 1.90  \\
            & TPMBM & 334.73 $\pm$ 9.21 & 97.58 & 235.68 & 0.12 & 1.35  \\ 
            \midrule
        \multirow{2}{*}{7}
            & D3AS & 367.31 ± 8.98 & 204.54 & 156.18 & 3.36 & 3.23  \\
            & TPMBM & 462.71 $\pm$ 11.73 & 64.89 & 396.84 & 0.00 & 0.99  \\
            \midrule
        \multirow{2}{*}{8}
            & D3AS & 254.13 ± 6.62 & 146.78 & 103.43 & 2.51 & 1.41  \\
            & TPMBM & 322.05 $\pm$ 8.99 & 89.98 & 230.70 & 0.24 & 1.12  \\ 
            \midrule
        \multirow{2}{*}{9}
            & D3AS & 302.53 ± 7.88 & 157.59 & 139.28 & 3.52 & 2.13  \\
            & TPMBM & 416.98 ± 11.29 & 51.15 & 365.17 & 0.00 & 0.67  \\ 
            \midrule
        \multirow{2}{*}{10}
            & D3AS & 418.08 ± 9.53 & 195.40 & 216.40 & 2.66 & 3.62  \\
            & TPMBM & 540.06 ± 12.63 & 15.92 & 523.88 & 0.00 & 0.25  \\
        \bottomrule
    \end{tabular}
    \\[-10pt]
\end{table}

From the table, we can see that both algorithms perform similarly in tasks with low clutter intensity $\lambda_c$ and high detection probability $p_d$ (tasks 1-3), with D3AS being slightly better in terms of TGOSPA. The low clutter intensity reduces the number of total hypotheses that TPMBM has to keep track of, effectively making it easier for the tracker to estimate the multi-object posterior density. At the same time, the high detection probability decreases the number of hypotheses with non-negligible likelihood within the total hypothesis set, as misdetection associations become unlikely. This, in effect, makes the pruning used by TPMBM to maintain computational tractability not hurt its performance so much, as only a few hypotheses are actually needed to accurately represent the multi-object posterior density. 

On the other hand, as we decrease the detection probability and/or increase the clutter intensity, the total number of hypotheses (and the number of these that have non-negligible likelihoods) increases sharply (tasks 4-10). Furthermore, increasing the measurement noise reduces the ability of the model to remove unlikely associations using gating. This increase also exacerbates the impact of the state dependency of the measurement model, making the Gaussian approximations made by TPMBM less accurate. These factors are reflected in a substantial degradation of tracking performance, with a clear trend of worsening as the tasks get more challenging.

In contrast, changing the detection probability and increasing clutter intensity do not have the same impact on the deep learning model, which seems better able to handle the added challenge. %We stress that exactly the same model of D3AS is trained for each task without changing model capacity in any way (i.e., same number of parameters). 
As opposed to TPMBM, D3AS does not rely on Gaussian approximations and requires no heuristics; its computational complexity is quadratic on the number of time-steps and independent of the multi-object model's parameters. Although its TGOSPA performance does decrease as the tasks get more challenging, it does so at a slower rate than TPMBM, leading to an increasing performance gap between these algorithms. 

Looking at the TGOSPA decompositions for each task, we observe that in all tasks, D3AS obtains a lower miss rate than TPMBM while not increasing the false rate by a similar amount. This indicates that D3AS is better able to find difficult trajectories than TPMBM, especially in the harder tasks. In fact, for certain samples in task 10, TPMBM was not able to predict any trajectories (all of the Bernoulli components had their existence probabilities below the extraction threshold). Moreover, TPMBM obtained considerably lower localization costs than D3AS for the harder tasks. However, we attribute this to TPMBM's high missed rate: TGOSPA's localization costs are only computed for trajectories that the tracker correctly detected, and results are not normalized by the number of detections. Normalizing by the number of detections does not change this, however, as TPMBM focused more on predictions for easier-to-identify trajectories in the harder tasks.

In addition to computing the TGOSPA scores on all tasks, we also analyze the performance in each subtask of data association and smoothing independently.
We use the TAA performance measure, as defined in Sec. \ref{subsec:taa}, to analyze the performance of D3AS and TPMBM in the data association subtask. TPMBM's data association is extracted from the association histories of each track in the most likely global hypothesis, whereas D3AS's data association comes directly from the output of the DDA module. The TAA for each task is shown at the top of Table \ref{tab:decomposition_scores} (higher values are better).
\begin{table*}[]
\centering
\caption{Top-1 association accuracy and TGOSPA with ground-truth associations for all tasks.  \label{tab:decomposition_scores}}
\resizebox{\textwidth}{!}{%
\begin{tabular}{@{}cccccccccccc@{}}
    \toprule
    \textbf{Evaluation}     &
    \textbf{Algorithm} &
    \textbf{Task 1}  &  \textbf{Task 2}  &  \textbf{Task 3}  &
    \textbf{Task 4}  &  \textbf{Task 5}  &  \textbf{Task 6}  &
    \textbf{Task 7}  &  \textbf{Task 8}   & \textbf{Task 9}   & \textbf{Task 10}
    \\
    \midrule
    \multirow{2}{*}{Top-1 A.A.}  
    & DDA  
        & 0.971 ± 0.002
        & 0.965 ± 0.003
        & 0.931 ± 0.004
        & 0.939 ± 0.003
        & 0.867 ± 0.006
        & 0.799 ± 0.007
        & 0.661 ± 0.009
        & 0.811 ± 0.008
        & 0.717 ± 0.008
        & 0.468 ± 0.010  \\
    & TPMBM    
        & 0.966 ± 0.003 
        & 0.951 ± 0.004
        & 0.889 ± 0.006
        & 0.884 ± 0.005
        & 0.765 ± 0.009
        & 0.605 ± 0.011 
        & 0.336 ± 0.012
        & 0.617 ± 0.010
        & 0.409 ± 0.011
        & 0.102 ± 0.008  \\
    \midrule
    \multirow{2}{*}{TGOSPA} 
    & DS  
        & 93.90 ± 2.58
        & 94.84 ± 2.61
        & 120.94 ± 3.17
        & 99.08 ± 2.55
        & 166.94 ± 4.53
        & 180.88 ± 4.56
        & 238.71 ± 6.14        
        & 166.44 ± 4.18
        & 180.29 ± 4.87         
        & 257.86 ± 6.38           \\
    & TPMBM    
        &  133.60 ± 3.85
        &  136.16 ± 4.00
        &  172.19 ± 4.85
        &  134.35 ± 3.89
        &  201.17 ± 5.62             
        &  210.16 ± 5.55  
        &  263.62 ± 6.86        
        &  200.91 ± 5.35          
        &  204.61 ± 5.70       
        &  257.30 ± 6.67          \\ 
    \bottomrule
\end{tabular}}
\vspace{-10pt}
\end{table*}
The results show that both algorithms can produce near-perfect data associations for the simpler tasks, but association accuracy drops as the tasks get harder. We see a similar trend in this table, showing that the performance gap between the algorithms increases for harder tasks.
\\[-10pt]

We can also evaluate performance on the smoothing task only. To do so, we feed both algorithms the ground-truth data associations for each measurement, which relieves them from needing to estimate the correct partitions for each track. Hence, all the tracks fed to the algorithms' smoothing modules only contain the true measurements from ea ch object in the scene. Doing so transforms the MOT task into the single-object setting without any false measurements, effectively evaluating only the smoothing performance of D3AS and TPMBM. We evaluate both methods in this setting using the sum of TGOSPAs between each predicted trajectory and the ground-truth trajectory of its corresponding measurement sequence. The results are shown in the bottom of Table \ref{tab:decomposition_scores} (lower values are better). 
\\[-10pt]

Here we see that smoothing performance is superior for the DS module in almost all tasks. Upon closer inspection of individual samples, we found indication that TPMBM can estimate trajectories better than D3AS when they are close to the sensor but becomes considerably worse as trajectories get closer to the edges of the field-of-view (where the measurement model becomes more nonlinear). Among other failure modes, TPMBM incorrectly estimates the start/end times in many of the trajectories, incurring high costs in terms of TGOSPA. 
\\[-10pt]

Regarding the performance gap between algorithms, the situation is now different than for the TAA performance measure: as the tasks become more complicated, the performance gap between D3AS and TPMBM decreases. We see two main reasons for this behavior. First, increasing the measurement noise, in specific $\sigma_{r,\text{min}}^2$ and $\sigma_{\theta, \text{min}}^2$, has the added effect of making the measurement noise covariance $\boldsymbol\Sigma(\mathbf x)$ less state-dependent, since $\sigma_{r,\text{max}}^2$ and $\sigma_{\theta, \text{max}}$ are kept fixed for all tasks, c.f. \eqref{eq:meas_noise_cov}. Second, when feeding each measurement sequence (partitioned using ground-truth associations) to the DS module, the confidences for each measurement are set to $1$. This essentially evaluates the model under out-of-distribution samples, especially for the harder tasks where the DDA module rarely predicted measurements with high confidences. Ideally, the DS module should be retrained for this setting, but computational limits prevented us from doing so. Nevertheless, we provide these results as they show that in the harder tasks the main advantage of D3AS is being able to perform data association better, therefore validating the efficacy of the DDA module.

% \section*{APPENDIX}
% \begin{appendices}
% \section{First appendix}
% \label{appx:first_one}
% First one here.

% \section{Second appendix}
% \label{appx:second_one}
% Second one here.

% \end{appendices}

\bibliographystyle{IEEEtran}
\bibliography{refs}

\end{document}